\documentclass[final]{cvpr}
\usepackage{tikz}
\usetikzlibrary{calc,spy}
\usepackage{graphicx}
\usepackage{comment}
\usepackage{amsmath,amssymb} 
\usepackage{xcolor}
\usepackage{wrapfig}
\usepackage{subcaption}
\usepackage{multicol}
\usepackage{multirow}
\usepackage[pagebackref=true,breaklinks=true,colorlinks,bookmarks=false]{hyperref}
\usepackage[british,abbreviations]{foreign}
\usepackage{array}
\usepackage{tabularx}
\usepackage{stackengine}
\usepackage{graphicx}
\usepackage{algorithm}
\usepackage{algorithmic}
\usepackage{amsthm}
\usepackage[toc,page]{appendix}
\usepackage{upgreek}

\newcommand{\curv}[1]{\breve{#1}}
\newtheorem{theorem}{Theorem}
\newtheorem{property}{\sc Property}
\newtheorem{definition}{\sc Definition}

\graphicspath{{fig/}}

\begin{document}

\def\cvprPaperID{2574} 
\def\confYear{CVPR 2021}

\title{Confluent Vessel Trees with Accurate Bifurcations}
\author{Zhongwen Zhang$^1$ \hspace{2ex} Dmitrii Marin$^{1,2}$ \hspace{2ex}  
Maria Drangova$^3$ \hspace{2ex} Yuri Boykov$^{1,2}$  \hspace{2ex}   \\[0.5ex]
{\small $^1$University of Waterloo, Canada \hspace{2ex}  $^2$Vector Research Institute, Canada  \hspace{2ex}  $^3$Robarts Research, Canada}  }

\maketitle
\thispagestyle{empty}
\pagestyle{empty}
\begin{abstract}
We are interested in unsupervised reconstruction of complex near-capillary vasculature with thousands of bifurcations where 
supervision and learning are infeasible. 
Unsupervised methods can use many structural constraints, \eg topology, geometry, physics.
Common techniques use variants of MST on geodesic {\em tubular graphs} minimizing symmetric pairwise costs, \ie distances. 
We show limitations of such standard undirected tubular graphs producing typical errors at bifurcations where flow ``directedness'' is critical. 
We introduce a new general concept of {\em confluence} for continuous oriented curves forming vessel trees and show how to 
enforce it on discrete tubular graphs. While confluence is a high-order property, we present an efficient practical algorithm 
for reconstructing confluent vessel trees using {\em minimum arborescence} on a directed graph enforcing confluence via 
simple flow-extrapolating arc construction. 
Empirical tests on large near-capillary sub-voxel vasculature volumes demonstrate significantly improved reconstruction accuracy at bifurcations. Our code has also been made publicly available \footnote{\href{https://vision.cs.uwaterloo.ca/code/}{https://vision.cs.uwaterloo.ca/code.}}.  
\end{abstract}

\section{Introduction}

\begin{figure}[h]
    \setlength{\fboxsep}{1pt}
    \setlength{\tabcolsep}{1pt}
    \centering
        \begin{tabular}{ccc}
            \includegraphics[height=0.4\linewidth]{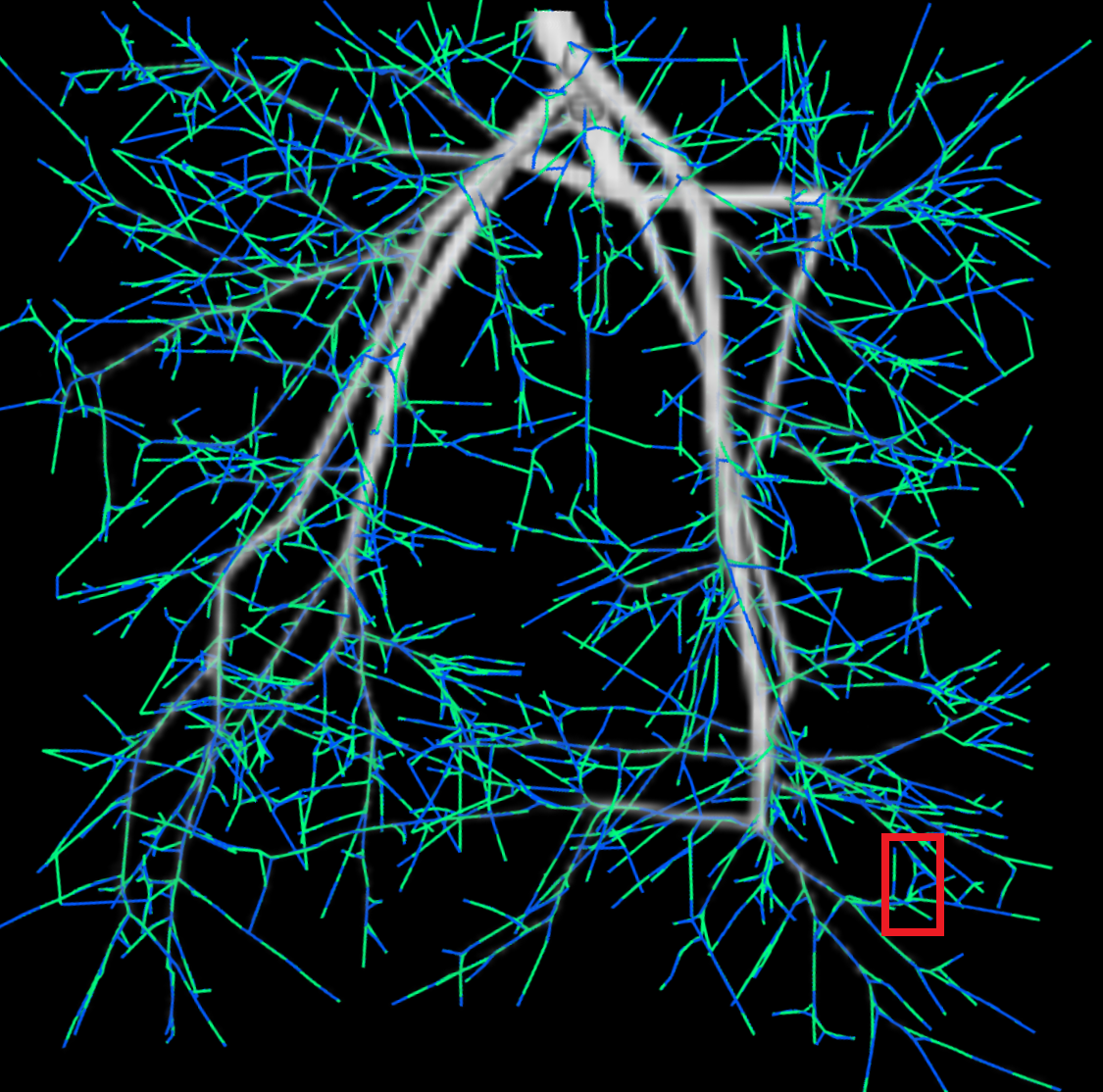} &
            \colorbox{red}{\includegraphics[height=0.28\linewidth,width=0.4\linewidth,angle=90]{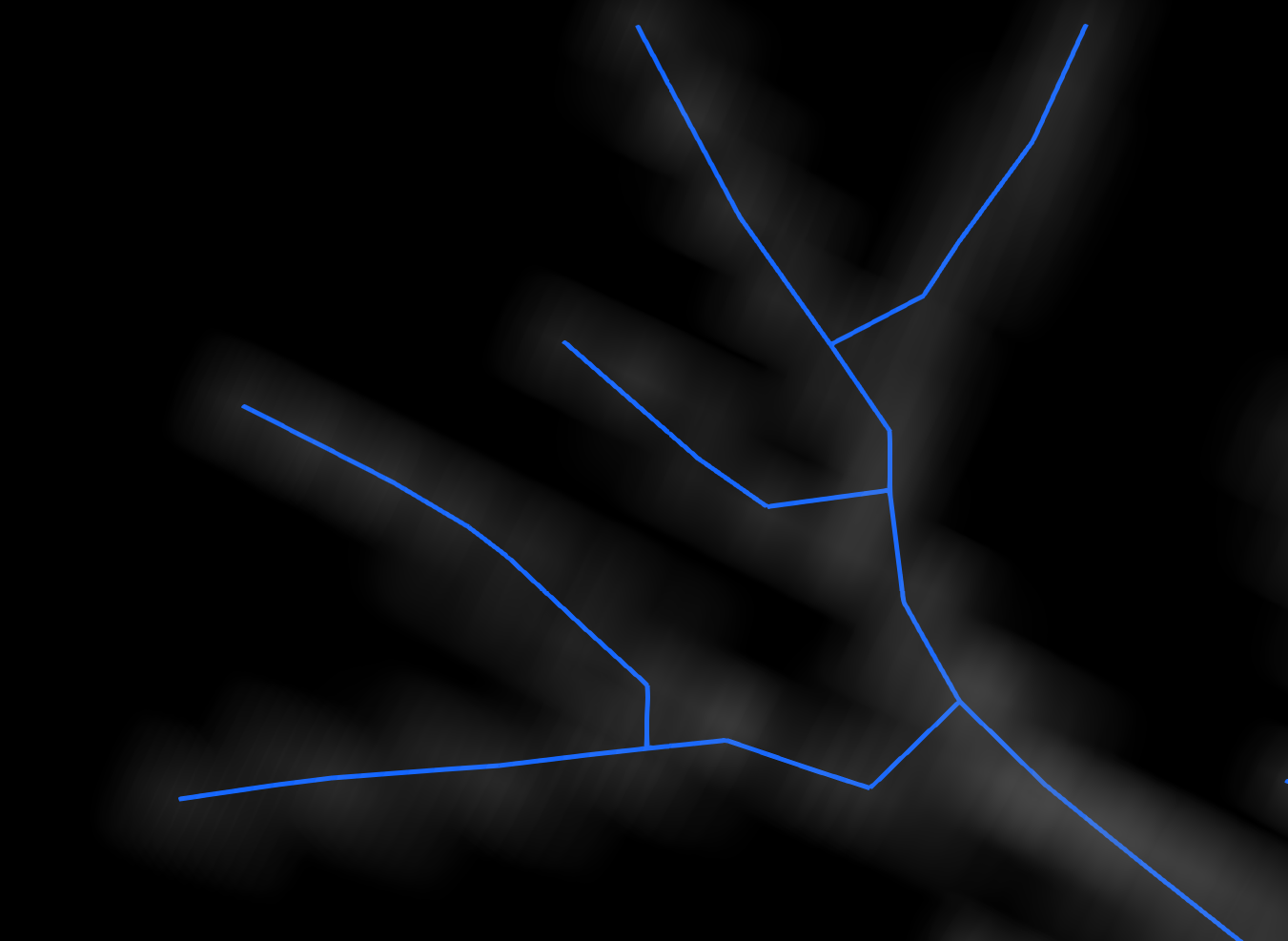}} &
             \colorbox{red}{\includegraphics[height=0.28\linewidth,width=0.4\linewidth,angle=90]{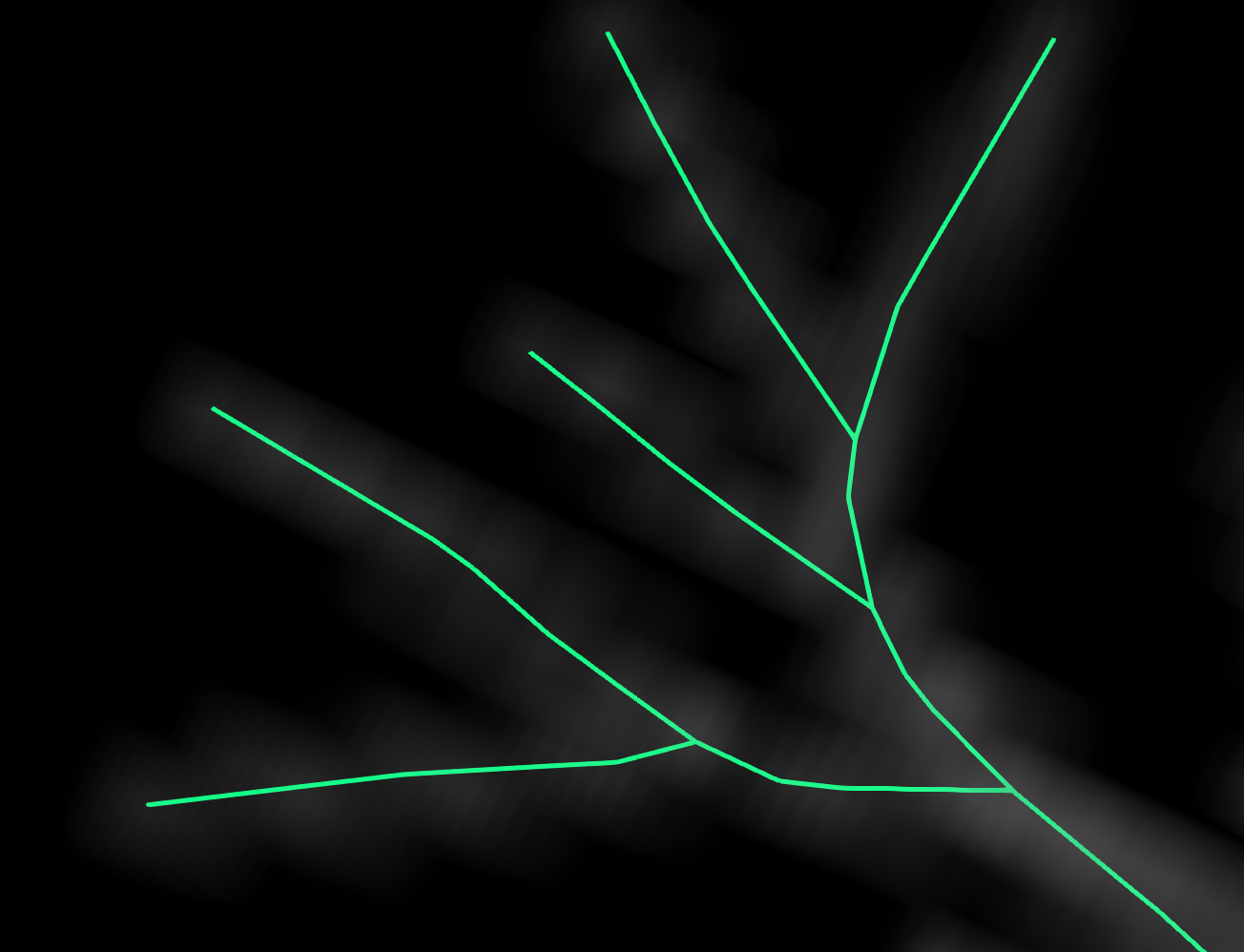}} \\
            \parbox{0.4\linewidth}{\scriptsize (a) synthetic raw data with two trees (blue \& green)} & \parbox{0.27\linewidth}{\scriptsize (b) geodesic graph MST \cite{gonzalez2008automated,xie2010automatic,turetken2013detecting,moriconi2018inference}} &
            \parbox{0.25\linewidth}{\scriptsize (c) confluent tree reconstruction }
        \end{tabular}
    \caption{{\em Synthetic example}: (a) raw 3D data with blue \& green reconstructed trees, see also zoom-ins (b,c). The blue tree (a,b) is an MST on a geodesic tubular graph. 
    The green tree (a,c), is a {\em minimum arborescence} on a directed {\em confluent tubular graph}, see \secref{sec:method}.
    \label{fig:teaserImages}}
\end{figure}

This paper is focused on unsupervised vessel tree estimation in large volumes containing numerous near-capillary vessels 
and thousands of bifurcations, see Figs.~\ref{fig:teaserImages}, \ref{fig:real data}. 
Around $80\%$ of the vessels in such data have sub-voxel diameter resulting in {\em partial volume} effects such as contrast loss and gaps. 
Besides the topological accuracy of trees reconstructed from such challenging imagery, 
we are particularly interested in the accurate estimation of bifurcations due to their importance in biomedical and pharmaceutical research. 

\subsection{Unsupervised vasculature estimation methods} 

Unsupervised vessel tree estimation methods for complex high-resolution volumetric vasculature data combine low-level vessel filtering and algorithms for computing global tree structures based on constraints from anatomy, geometry, physics, \etc. Below we review the most
relevant standard methodologies.

\begin{figure}
    \centering
    \setlength{\tabcolsep}{0ex}
    \begin{tabular}{m{0.45\linewidth}cm{0.45\linewidth}}
        \includegraphics[height=1\linewidth]{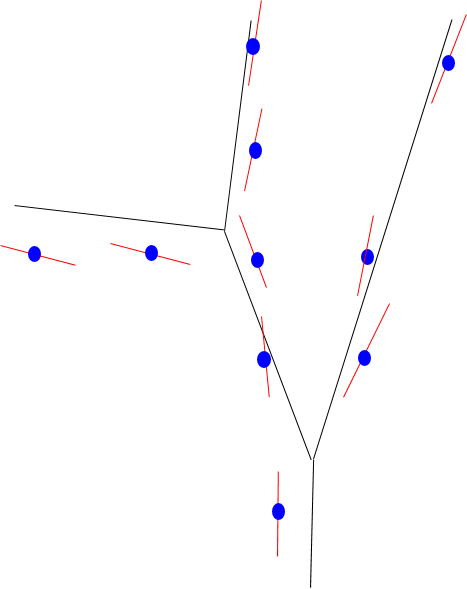} & \hspace{3ex} &
        \includegraphics[height=1\linewidth]{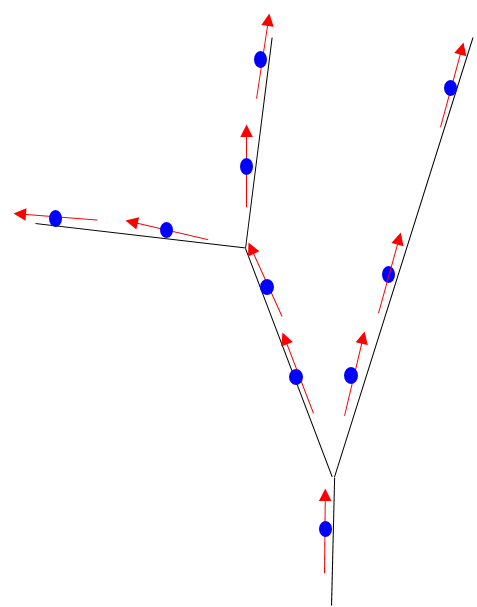} \\
     \centering\scriptsize (a) Frangi filtering \cite{frangi1998multiscale} && \centering\scriptsize (b) {\em oriented} flow pattern~\cite{Divergence:cvpr19}
    \end{tabular}
    \caption{{\em Low-level vessel estimation}: True centerline is black. 
    Blue voxels in (a) are local maxima of some tubularity measure \cite{frangi1998multiscale,law2008oof,turetken2013detecting,freeman1991design} 
    in the direction orthogonal to the estimated centerline tangents (red). 
    Regularization \cite{zucker:89,thin:iccv15} can estimate subpixel centerline points (b) 
    and oriented tangents \cite{Divergence:cvpr19} (red flow field).
    \label{fig:low_level_vessels}}
\end{figure}

{\bf Low-level vessel estimation:} Anisotropy of tubular structures is exploited by standard vessel filtering techniques, \eg Frangi \etal  \cite{frangi1998multiscale}.
Combined with non-maximum suppression, local tubularity filters provide estimates for vessel centerline points and tangents, see \figref{fig:low_level_vessels}(a). 
Technically, elongated structures can be detected using intensity Hessian spectrum \cite{frangi1998multiscale}, {\em optimally oriented flux} models \cite{law2008oof,turetken2013detecting}, steerable filters \cite{freeman1991design}, path operators \cite{merveille2017curvilinear} or other anisotropic models. Dense local vessel detections can be denoised using curvature regularization \cite{zucker:89,thin:iccv15}. 
Prior knowledge about divergence or convergence of the vessel tree (arteries vs veins) can also be exploited 
to estimate an {\em oriented} flow pattern \cite{Divergence:cvpr19}, see \figref{fig:low_level_vessels}(b). 

{\bf Thinning:}
One standard approach to vessel topology estimation is via {\em medial axis} \cite{siddiqi2008medial}.
This assumes known vessel segmentation (volumetric mask) \cite{merveille2019n}, which can be computed only for relatively thick vessels. 
Well-formulated segmentation of thin structures requires Gaussian- or min-curvature surface regularization that has no known
practical algorithms. Segmentation is particularly unrealistic for sub-voxel vessels.

{\bf Geodesics and shortest paths:}
Geodesics \cite{cohen:MIA01,chen2017global} and shortest paths \cite{figueiredo:TMI95} are often 
used for $AB$-interactive reconstruction of vessels between two specified points. 
A vessel is represented by the shortest path with respect to some anisotropic continuous (Riemannian) 
or discrete (graph) metric based on a local tubularity measure.  
Interestingly, the minimum path in an ``elevated" search space combining spatial locations and radii can simultaneously 
estimate the vessel's centerline and diameter, implicitly representing vessel segmentation \cite{li2007vessels,benmansour2011tubular}. 
Unsupervised methods widely use geodesics as their building blocks.

\textbf{Spanning trees:}
The standard graph concept of a {\em minimum spanning tree} (MST) is well suited for unsupervised reconstruction of large trees 
with unknown complex topology \cite{gonzalez2008automated,xie2010automatic,turetken2013detecting,moriconi2018inference}. 
MST is closely related to the {\em shortest paths} and {\em geodesics} since its optimality is defined with respect to its length. 
Like shortest paths, globally optimal MST can be computed very efficiently. In contrast to the shortest paths, 
MST can reconstruct arbitrarily complex trees without user interaction. 

\begin{figure}
    \centering
    \setlength{\tabcolsep}{0ex}
    \begin{tabular}{m{0.45\linewidth}cm{0.45\linewidth}}
        \includegraphics[height=1\linewidth]{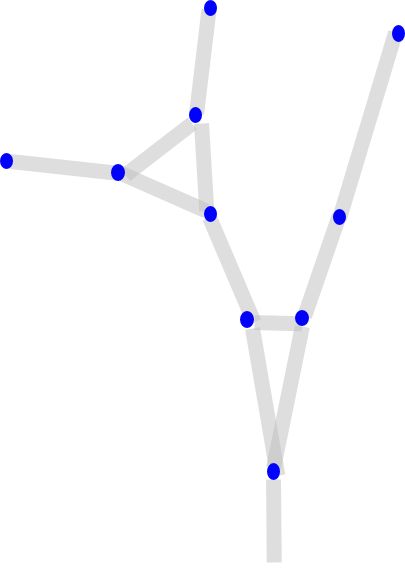} & \hspace{3ex} &
        \includegraphics[height=1\linewidth]{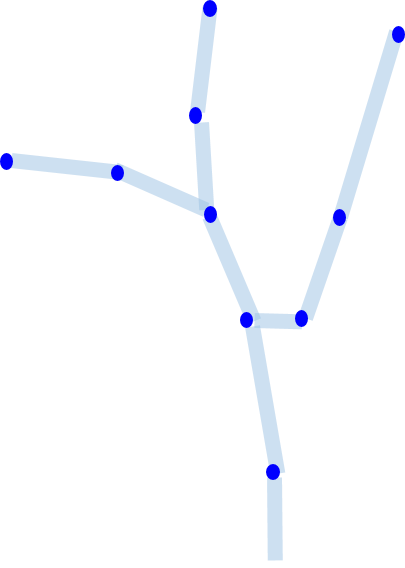} \\
        \centering\scriptsize (a) geodesic tubular graph && 
        \centering\scriptsize (b) MST
    \end{tabular}
    \caption{{\em Global vessel tree reconstruction}: (a) geodesic tubular graph is based on low-level estimates 
    in \figref{fig:low_level_vessels}.
    Graph edges represent distances, geodesics, or other symmetric (undirected) properties. 
    MST reconstruction quality (b) depends on the graph construction (nodes, neighborhoods, edge weights).
    \label{fig:global_vessels}}
\end{figure}

\begin{figure*}[t]
    \centering
    \setlength{\tabcolsep}{1pt}
    \begin{tabular}{ccccc}
        \includegraphics[height=0.3\linewidth]{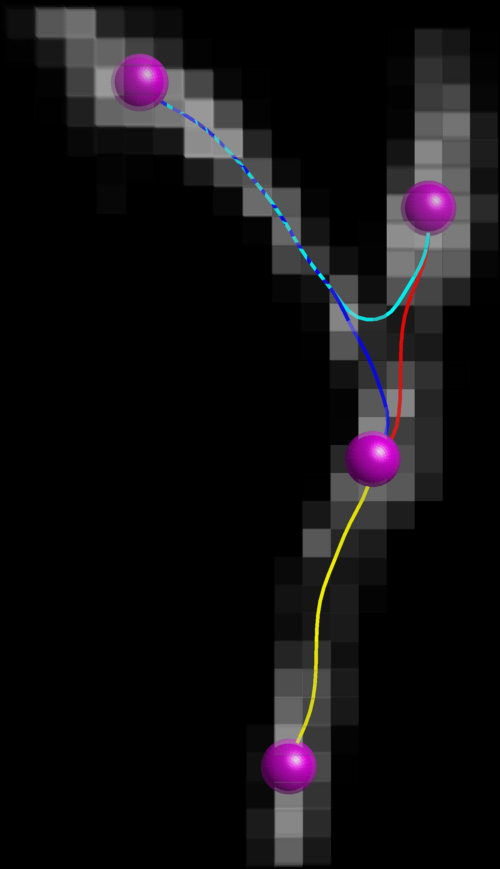} &&
        \includegraphics[height=0.3\linewidth,width=0.24\linewidth]{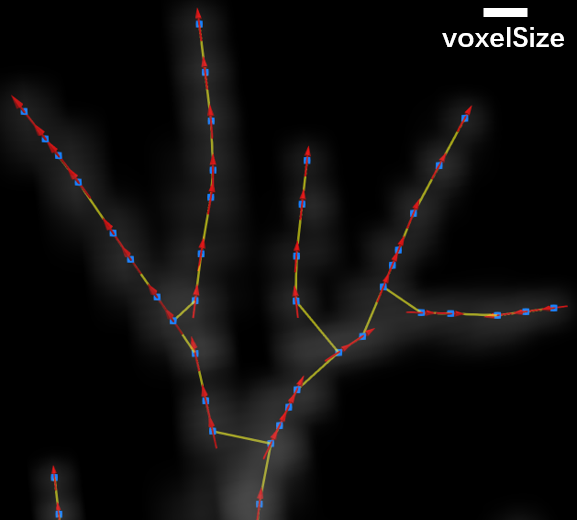} &
        \includegraphics[height=0.3\linewidth,width=0.24\linewidth]{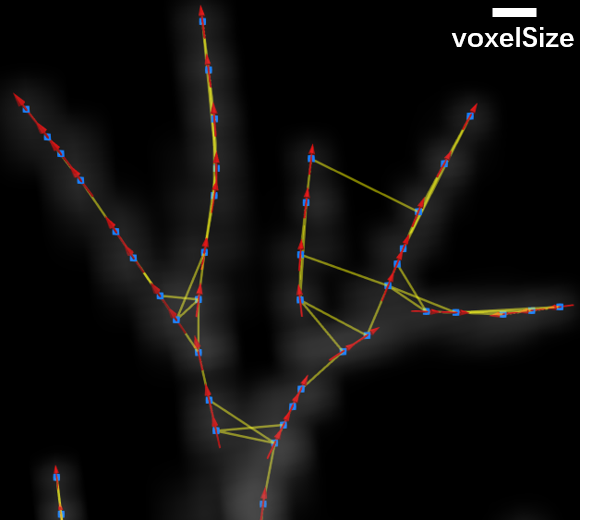} &
        \includegraphics[height=0.3\linewidth,width=0.24\linewidth]{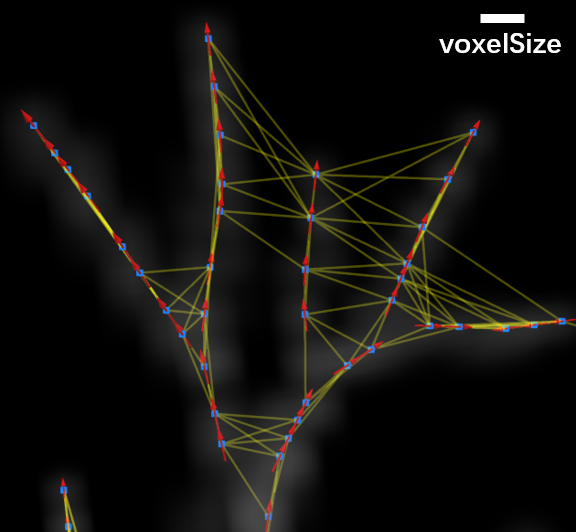} \\[-5mm]
        \tiny \color{white} image from \cite{turetken:Neuroinform2011} && \hfill \color{white} $k=2$~ & \hfill \color{white} $k=3$~ & \hfill \color{white} $k=6$~ \\
        (a) sparse graph & \hspace{4mm} & \multicolumn{3}{c}{(b) semi-dense geodesic tubular graphs with different size NN systems}
    \end{tabular}
    \caption{Examples of standard {\em geodesic tubular graphs} for vascular image data: A graph from \cite{turetken:Neuroinform2011} in (a) uses (purple) nodes connected by undirected edges corresponding to
    shortest paths (geodesics) w.r.t. tubularity-based Riemanninian metric. Alternatively, (b) shows graphs where (blue) nodes correspond to densely-sampled voxels along the vessels \cite{Divergence:cvpr19}. Here the edges correspond to
    nearest neighbors (KNN) weighted by length of some spline interpolation. 
    In both cases (a) and (b), near- or sub-voxel vessels have sparsely sampled bifurcations.
    \label{fig:tubular graph illustration}}
\end{figure*}

The quality of MST vessel tree reconstruction depends on the underlying graph construction, 
see Figs.~\ref{fig:global_vessels} and \ref{fig:gridarb}(a).
Graphs designed for reconstructing thin tubular structures as their spanning tree (or sub-tree) are often called {\em tubular graphs}.
Typically, the nodes are ``anchor'' points generated by low-level vessel estimators, \eg see \figref{fig:low_level_vessels}.
Such anchors represent sparse \cite{turetken:PAMI2016} or semi-dense \cite{thin:iccv15} samples from the estimated tree structure 
that may be corrupted by noise and outliers.
Pairwise edges on a tubular graph typically represent distances or geodesics between the nodes, 
as in $AB$-interactive methods discussed earlier. Such graphs are called {\em geodesic tubular graphs}, see \figref{fig:tubular graph illustration}.

There are numerous variants of tubular graph constructions designed to represent various thin structures as MST  \cite{gonzalez2008automated,xie2010automatic,turetken2013detecting,moriconi2018inference}
or shortest path trees \cite{peng2011automatic}. There are also interesting and useful extensions of MST addressing tubular graph outliers, \eg
k-MST \cite{turetken:Neuroinform2011} and integer programming technique in \cite{turetken:PAMI2016}. Such approaches are more powerful as they seek
minimum sub-trees that can automatically exclude outliers. However, the corresponding optimization problems are NP-hard and require approximations. 
Such methods are expensive compared to the low-order polynomial complexity of MST. They are not practical for dense
reconstruction problems in high-resolution vasculature volumes.

\subsection{Motivation and contributions}
We are interested in unsupervised reconstruction of large complex trees from vasculature volumes resolving 
near-capillary details. Common geodesic approaches can not represent asymmetric smoothness at bifurcations, which
have forms sensitive to flow orientation. Hence, standard methods produce vessel tree reconstructions with 
significant bifurcation artifacts, see Figs.~\ref{fig:teaserImages}(b), \ref{fig:global_vessels}(b) and \ref{fig:gridarb}(a).  
We define a general geometric property for oriented vessels, {\em confluence}, which is missing in prior art,
and propose a practical graph-based reconstruction method enforcing it. The reconstructed
confluent vessel trees have significantly better bifurcation accuracy. Our contributions are detailed below.

{ \renewcommand{\item}{$\bullet$ }

\item We introduce {\em confluence} as a geometric property for overlapping oriented smooth curves in $\mathbb R^3$, \eg representing
blood-flow trajectories\footnote{Confluence is known in other contexts, \eg rail tracks \cite{hui2007train}.}\@. It is like
``co-differentiability'' or ``co-continuity''. We define confluent vessel trees formed by overlapping oriented curves. 

\item We extend confluence to discrete paths and trees on directed tubular graphs where
directed arcs/edges represent continuous oriented arcs/curves in $\mathbb R^3$.
We propose a simple {\em flow-extrapolating circular arc} construction that guarantees $\varepsilon$-confluence, which approximates confluence.
Our confluence constraint implies {\em directed} tubular graph with asymmetric edge weights, 
which is in contrast to standard {\em undirected} geodesic tubular graphs \cite{jomier2005automatic, gonzalez2008automated, xie2010automatic, turetken2013detecting,moriconi2018inference, turetken:Neuroinform2011, thin:iccv15, turetken:PAMI2016, Divergence:cvpr19}.

\item We present an efficient practical algorithm for reconstructing {\em confluent vessel trees}. It uses
{\em minimum arborescence} \cite{edmonds:67,tarjan:77} on our directed {\em confluent tubular graph} construction.

\item Our experiments on synthetic and real data confirm that confluent tree reconstruction significantly 
improves bifurcation accuracy. We demonstrate qualitative and quantitative improvements via standard 
and new accuracy measures \footnote{For our dataset and implementation of evaluation metrics discussed in this paper see \href{https://vision.cs.uwaterloo.ca/data/}{https://vision.cs.uwaterloo.ca/data}.} evaluating tree structure, bifurcation localization, and bifurcation angles. 

}

Our concept of confluent trees is general and our specific algorithm can be modified or extended in many ways, 
some of which are discussed in \secref{sec:method}.
To explicitly address outliers, minimum arborescence on our confluent tubular graph can be replaced by 
optimal sub-tree algorithms \cite{turetken:Neuroinform2011,turetken:PAMI2016}\footnote{IP solver in \cite{turetken:PAMI2016}
uses {\em minimum arborescence} as a subroutine.} or explicit outlier detection \cite{thin:iccv15}, 
but these approximation algorithms address NP-hard problems and maybe too expensive for large semi-dense tubular graphs we study in this work. 
While outlier detection is relevant, this work is not focused on this problem.

\section{Confluence of Oriented Curves}

\begin{figure}
    \centering
    \usetikzlibrary{decorations.pathreplacing,decorations.markings}
    
    \setlength{\tabcolsep}{0pt}
    \begin{tabular}{cc}
    \begin{tikzpicture}[
            ]
        \coordinate (A) at (0,0);
        \coordinate (B) at (0,2);
        \coordinate (P) at (2,1);
        \coordinate (C) at (3.7,2);
        \coordinate (C2) at (3,0.5);
        \coordinate (D) at (4,1.5);
        
        \node[below] at (P) {$p$};
        
        \begin{scope}[very thick,decoration={
                markings,
                mark=at position 0.25 with {\arrow{>}}}
            ] 
            \draw [ultra thick,red,postaction={decorate}] (A) to[out=0,in=180] node[label= below:$\alpha$] {} (P) to[out=0,in=190] (C2) to[out=10,in=-90] (C);
            \draw [ultra thick,blue,postaction={decorate}] (B) to[out=0,in=180] node[label= below:$\beta$] {} (P) to[out=0,in=190] (D);
        \end{scope}
        \draw[fill] (P) circle (0.05);
    \end{tikzpicture} &
    \begin{tikzpicture}
        \coordinate (A) at (0,0);
        \coordinate (A2) at (0,0.03);
        \coordinate (B) at (0,2);
        \coordinate (P) at (2,1);
        \coordinate (P2) at (2,1.03);
        \coordinate (C) at (4,0.5);
        \coordinate (D) at (4,1.5);
        
        \begin{scope}[very thick,decoration={
                markings,
                mark=at position 0.75 with {\arrow{>}}}
            ] 
            \draw [ultra thick,red,postaction={decorate}] (A) to[out=0,in=180] (P) to[out=0,in=190] node[label= below right:$\alpha$] {} (C);
            
            \draw [ultra thick,blue,postaction={decorate}] (A2) to[out=0,in=180] (P2) to[out=0,in=190] node[label= above right:$\beta$] {} (D);
        \end{scope}
    \end{tikzpicture} \\
    (a) confluence at point $p$ & (b) confluent curves
    \end{tabular}
    \caption{Confluence for oriented curves $\alpha$ and $\beta$.} \label{fig:confluence illustration}
    \label{fig:my_label}
\end{figure}
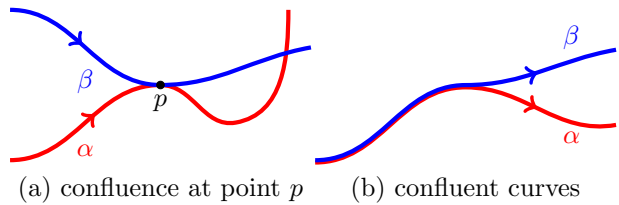

This section introduces geometrically-motivated concept of smoothness for objects containing multiple oriented curves,
such as vessel trees. We define {\em confluence} as follows.
\begin{definition}[confluence at a point] \label{def:confluence of curves}
Two differentiable oriented curves $\alpha(t)$ and $\beta(\tau)$ 
are called confluent at a shared point $p$ if for some $k>0$
$$
\alpha'(t_p) = k \, \beta'(\tau_p) \quad \text{where}
$$
$t_p$ and $\tau_p$ are s.t. $\alpha(t_p) = \beta(\tau_p) = p$, see \figref{fig:confluence illustration}(a).
\end{definition}
We will call two oriented curves {\em confluent} if they are confluent at all points they share, see \figref{fig:confluence illustration}(b). 

\begin{figure*}[t]
    \centering
    \begin{tabular}{ccc}
        \includegraphics[scale=0.25]{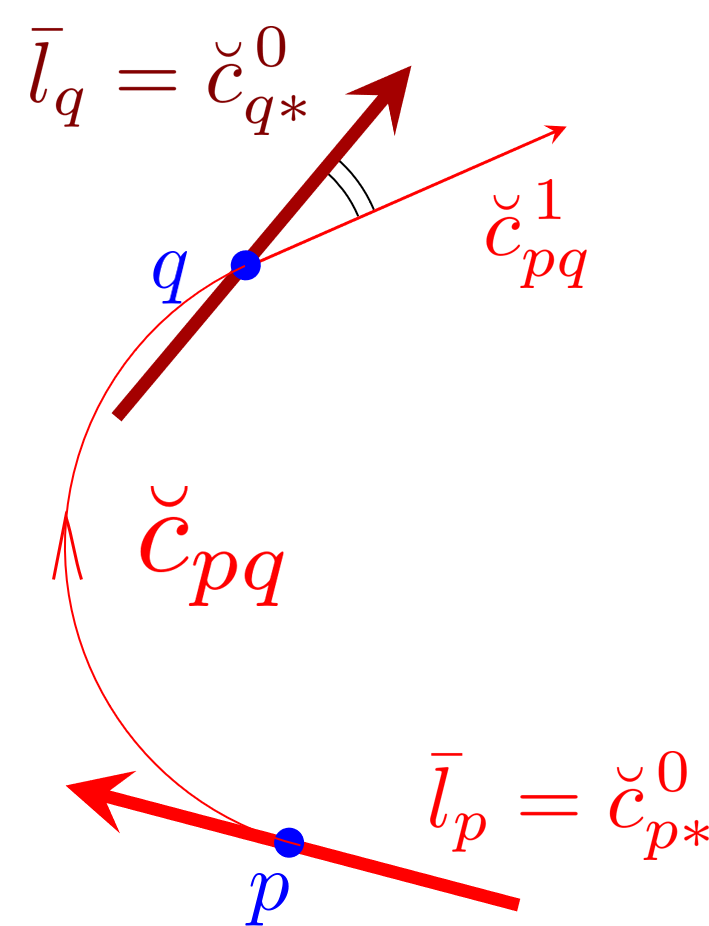} &
        ~~~~~\includegraphics[scale=0.25]{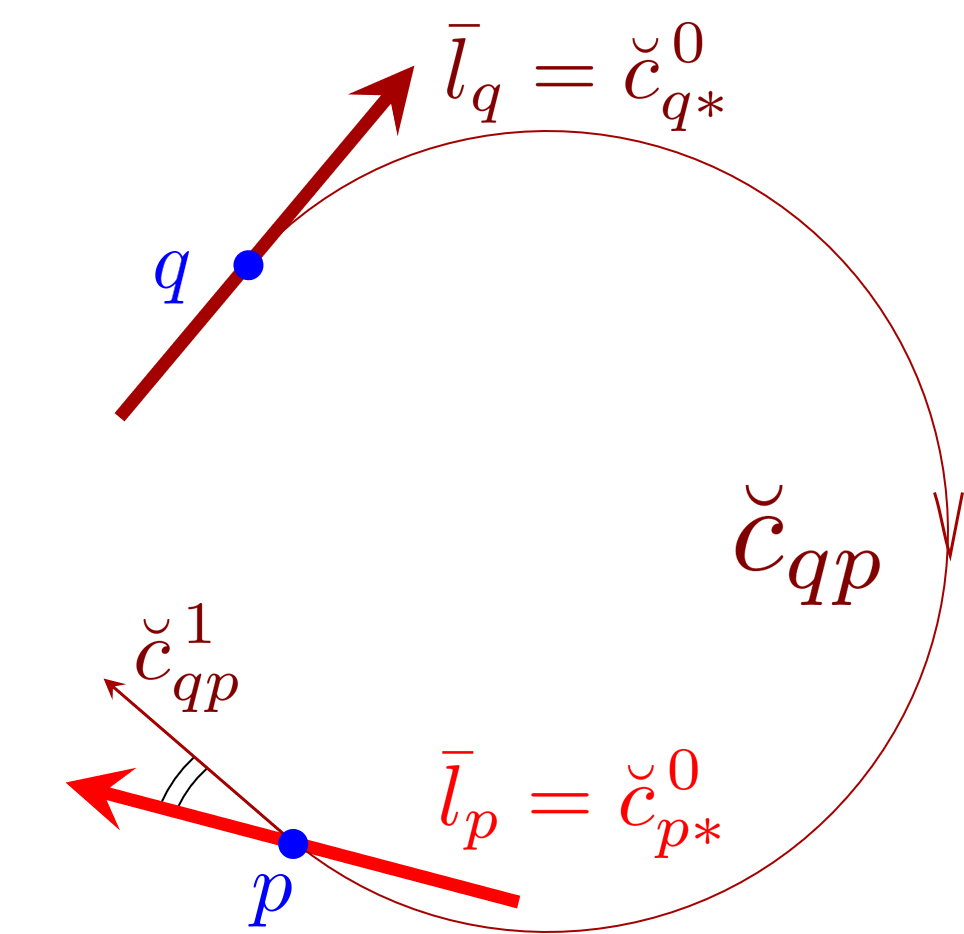}~~~~~ &
        \includegraphics[scale=0.25]{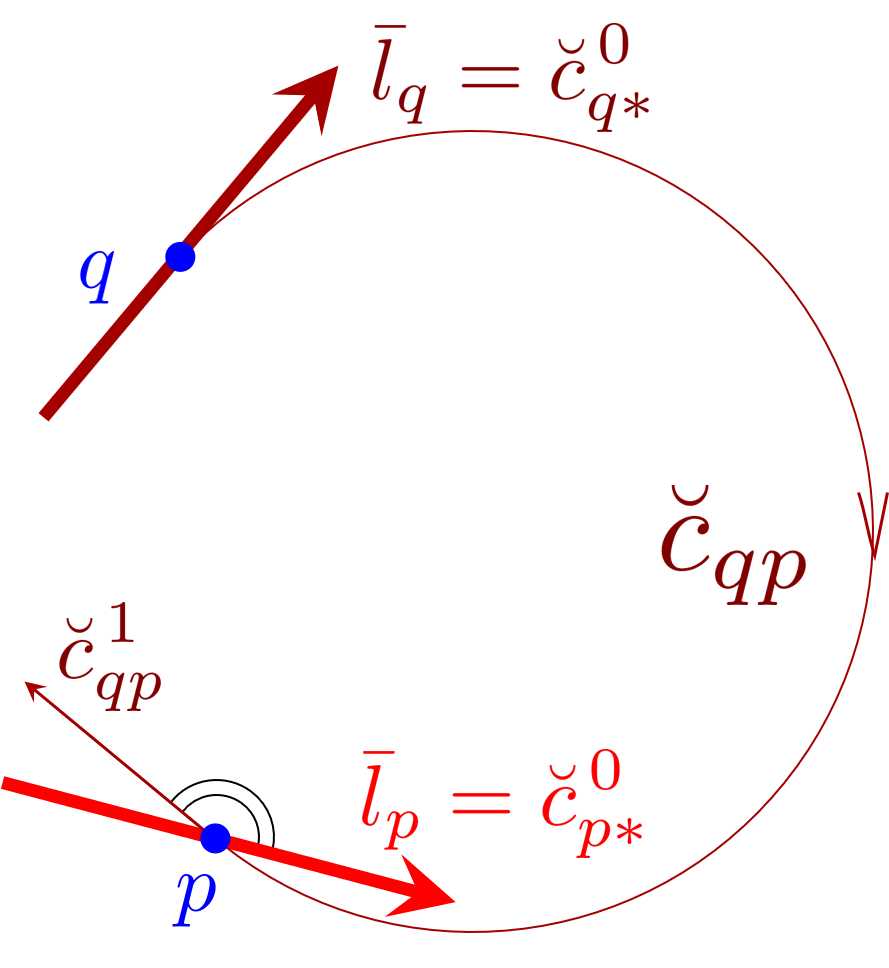} \\
        (a) $\;\angle (\curv c_{pq}^{\,1}, \curv c_{q*}^{\,0} ) \leq \varepsilon $ & 
        (b) $\;\angle (\curv c_{qp}^{\,1}, \curv c_{p*}^{\,0} ) \leq \varepsilon $ &
        (c) $\;\angle (\curv c_{qp}^{\,1}, \curv c_{p*}^{\,0} ) \leq \varepsilon $ \\
         $\;\;\;\;$ short confluent arc $\curv c_{pq}$ & 
         $\;\;\;\;$ long confluent arc $\curv c_{qp}$ & 
         $\;\;\;\;$ non-confluent arc $\curv c_{qp}$
    \end{tabular}
    \caption{Examples of directed flow-extrapolating circular arcs: (a) flow $\breve c_{pq}^{\,1}$ extrapolated by 
    {\em confluent} arc $\breve c_{pq}$ from $p$ to $q$  is consistent with the local flow estimate 
    $\bar l_q =   \breve c_{q*}^{\,0}$ at point $q$ as the angle between two vectors is small. 
    Flow extrapolation from $q$ to $p$ in (b) requires another circular arc $\breve c_{qp}$ that belongs to 
    a different circle defined by tangent $\bar l_q$. Two arcs $\breve c_{pq}$ and $\breve c_{qp}$ are not even co-planar
    if tangents $\bar l_p$ and $\bar l_q$ are not. The extrapolated flow $\breve c_{qp}^{\,1}$ is also consistent 
    with the local flow estimate $\bar l_p=\breve c_{p*}^{\,0}$ at $p$, so that the reverse arc $\breve c_{qp}$ in (b) is confluent as well. 
    In (c) the local flow estimate $\bar l_p$ at $p$ 
    is flipped and arc $\breve c_{qp}$ becomes non-confluent since the angle between $\breve c_{qp}^{\,1}$ and 
    $\bar l_p = \breve c_{p*}^{\,0}$ is large. Note that arc $\breve c_{pq}$ in (c) differs from (a) but  it must be non-confluent as well, 
    see Theorem \ref{th:symmetry}.}
    \label{fig:directed weights}
\end{figure*}

Our concept of confluence is closely related to the geometric \textit{$\mathcal G^1$-continuity} \cite{derose1985geometric,fowler1966cubic}. 
A curve $\alpha$ is called $\mathcal G^1$-continuous if at any point on the curve the slope orientation is continuous.
Incidentally, the differentiability classes $C^k$ are too restrictive as a $\mathcal G^1$-continuous curve can easily be not $C^1$ due to the curve parameterization. Note that $\mathcal G^1$-continuity is only defined for a \textit{single} curve while our confluence extends it for a pair of curves and can be seen as ``co-$\mathcal G^1$-continuity''. 

Our concept of confluence allows defining arbitrarily complex (continuous) {\em confluent vessel trees}. Such trees are 
formed by multiple oriented curves representing motion trajectories of blood particles from the common root to an arbitrary number of leaves 
where each pair of curves must be confluent. \figref{fig:confluence illustration}(b) shows a simple example of a tree 
formed by two confluent curves with one bifurcation, which can be formally defined. 

\section{Confluent Tubular Graphs} \label{sec:method}

Our discrete approach to reconstructing {\em confluent vessel trees} is based on efficient algorithms for directed graphs. 
Our ``tubular'' graph nodes correspond to a finite set of detected vessel points. 
We use discrete representation of oriented vessels as paths along directed edges or {\em directed arcs} $(p,q)$ connecting the graph nodes. 
Each arc $(p,q)$ on our tubular graph represents
an oriented continuous ``flow-extrapolating'' curve in $\mathbb R^3$ from $p$ to $q$. Such curves could be obtained from physical models 
based on fluid dynamics. For simplicity, this paper is focused on oriented {\em circular arcs}, see \secref{sec:tubular_dir}, 
motivated as the lowest-order polynomial splines capable of enforcing $\mathcal G^1$-continuity and confluence.
In general, our confluent tubular graph construction can use higher-order flow-extrapolation models, \eg cubic Hermite splines that are 
common in computer graphics and geometric modeling of motion trajectories. 

By using circular {\em arcs} as flow-extrapolating curves, we introduce some ambiguity with ``arcs''  as the standard term for 
graph edges. However, this should not create confusion since there is a one-to-one relation between {\em directed arcs} 
on our tubular graph and the corresponding (circular) {\em oriented arcs} in $\mathbb R^3$. 
Note that both interpretations are oriented/directed. In all technically formal sentences, continuous or discrete 
interpretation of the ``arc'' is clear from the context. In more informal settings, both interpretations are often equally valid.

The rest of this Section is as follows. Oriented flow-extrapolating circular arcs between tubular graph nodes are introduced in 
\secref{sec:tubular_dir} where $\varepsilon$-confluence constraint is defined in the context of such arcs. We also define
directed arc weights to represent the confluence constraint and the local costs of sending flow along these arcs.
Geometric properties of confluent circular arcs are discussed in \secref{sec:cf-cs}. The algorithm estimating
confluent vessel trees via {\em minimum arborescence} on our directed tubular graph is presented in \secref{sec:alg}.

\subsection{Confluent flow-extrapolating arcs} \label{sec:tubular_dir}

Formally, tubular graph $G = \langle V,A \rangle$ is based on a set of nodes/points $V$ embedded in $\mathbb R^3$ representing semi-densely sampled centerlines of a tubular structure. 
$A \subseteq V^2$ is a set of directed arcs.
For our tubular graph construction, each directed arc $(p,q)\in A$ represents some continuous oriented curve in $\mathbb R^3$ modelling
flow-extrapolation from point $p$ to point $q$, see \figref{fig:directed weights}. As discussed earlier, this paper is focused on oriented circular arcs
as the simplest geometric model that can represent confluent vessels, even though higher-order geometric splines or 
physics-motivated curves are possible. Our specific construction uses flow-extrapolating circular arcs based on a set of 
unit vectors  $\bar L=\{\bar l_p\}_{p\in V} \subset S^2$ representing flow direction estimates at the nodes, 
see Fig.~\ref{fig:low_level_vessels}(b). 
Oriented circular arc $\curv c_{pq}$ is fit into starting point $p$, its flow orientation estimate $\bar l_p$, 
and the ending point $q$, see \figref{fig:directed weights}. 
Formally, curve $\curv c_{pq}$ corresponds to a differentiable function $$\curv c_{pq}\;: \;\; [0,1] \;\to\; \mathbb R^3$$ 
traversing points on a circular arc in the plane spanned by $p$, $q$, and vector $\bar l_p$ so that
\begin{equation} \label{eq:arc_def}
\curv c_{pq}(0) = p\;,\;\; \curv c_{pq}(1) = q\;,\;\; \frac{\curv c_{pq}'(0)}{\|\curv c_{pq}'(0)\|} = \bar l_p
\end{equation}
where derivative $\curv c_{pq}'(s)$ gives an oriented tangent.  


For shortness, we define (oriented) unit tangents at the beginning and the end points of any flow-extrapolating arc $\curv c$ as
\begin{equation}
    \curv c^{\,0} \; := \; \frac{\curv c'(0)}{\|\curv c'(0)\|} \;\;\;,\;\;\;\;\; \curv c^{\,1} \; :=\; \frac{\curv c'(1)}{\|\curv c'(1)\|}\;.
\end{equation}
The definition of arc $\curv c_{pq}$ in \eqref{eq:arc_def} implies $\curv c_{pq}^{\,0} \;\equiv\; \bar l_p $ so that
tangent $\curv c_{pq}^{\,0}$ is the same for any arc starting at given point $p$ regardless of its end point $q$. Thus,
\begin{equation} \label{eq:dot}
\curv c_{p*}^{\,0}\;\;\equiv\;\; \bar l_p
\end{equation}
where the star $*$ represents an arbitrary end point. On the other hand, tangent $\curv c_{pq}^{\,1}$ at the end point $q$
depends on the arc's starting point $p$. That is, generally,  
$$\angle (\curv c_{pq}^{\,1}, \curv c_{rq}^{\,1}) \;\;\not\equiv\;\; 0 \;\;\;\;\;\;\;\;\text{if}\;\;p\neq r.$$

\paragraph{$\varepsilon$-Confluence Constraint:} 
To constrain our tubular graph so that all feasible vessel trees are confluent, it suffices to
enforce confluence of the arcs at the nodes where they meet. However, our simple flow-extrapolating circular arcs \eqref{eq:arc_def}
can not be used to enforce the exact confluence.
We use some threshold $\varepsilon$ to introduce a relaxed version of confluence in 
Definition \ref{def:confluence of curves} for an arbitrary pair of adjacent arcs $\curv c_{pq}, \curv c_{qr}$ 
connecting points  $p$, $q$ and $r$
\begin{equation} \nonumber
     \angle (\curv c_{pq}^{\,1}, \curv c_{qr}^{\,0} )\;\leq\; \varepsilon.
\end{equation}
In general, this is a high-order (triple clique) constraint. But, property \eqref{eq:dot} of our 
flow-extrapolating arc construction shows that  the end point of the second arc $\curv c_{qr}$ is irrelevant. Indeed,
\begin{equation} \nonumber
     \angle (\curv c_{pq}^{\,1}, \curv c_{qr}^{\,0} ) \; =\;  \angle (\curv c_{pq}^{\,1}, \bar l_q )
     \;\equiv\; \angle (\curv c_{pq}^{\,1}, \curv c_{q*}^{\,0})
\end{equation}
implying that our specific tubular graph construction allows to express confluence as a pairwise constraint 
\begin{equation}
     \angle (\curv c_{pq}^{\,1}, \curv c_{q*}^{\,0} ) \;\leq\; \varepsilon
\end{equation}
for any pair of points $p$, $q$. In essence, this becomes a constraint for our flow extrapolating arcs $\curv c_{pq}$ that can be called
confluent if $\angle (\curv c_{pq}^{\,1}, \bar l_q ) \;\leq\; \varepsilon $, see \figref{fig:directed weights}.

To enforce $\varepsilon$-confluence constraint, our tubular graph can simply drop all non-confluent arcs. 
Thus, any directed vessel tree on our graph will be confluent by construction. 
This paper explores the simplest approach to reconstructing confluent vessel trees as the {\em minimum arborescence} 
on our directed tubular graph. In this case, instead of dropping non-confluent arcs, one can
incorporate $\varepsilon$-confluence constraint directly into the cost of the corresponding directed graph arcs
\begin{equation}\label{eq:assymetric weight}
    w_{pq} \;\;\; := \;\;\; 
        \begin{cases}
            \operatorname{length}(\curv c_{pq})& \text{if}\;\; \angle ( \curv c_{pq}^{\,1}, \curv c_{q*}^{\,0} ) \leq \varepsilon \\
            \;\;\;\infty             &\;\;\; \text{otherwise.}
        \end{cases}
\end{equation}
The reverse edge on our tubular graph has different weight $w_{qp}\neq w_{pq}$ because it corresponds to 
a different flow extrapolating arc $\curv c_{qp}$ that has a different length, see \figref{fig:directed weights}(b). 
As an extension, our approach also allows ``elastic'' arc weights by adding integral of arc's 
curvature to its length in \eqref{eq:assymetric weight}.
It is also possible to impose soft penalties for the discrepancy between the extrapolated flow $\breve c_{pq}^{\,1}$ and flow estimate 
$\breve c_{q*}^{\,0} \equiv \l_q $ in \eqref{eq:assymetric weight} based on physical, physiological, or other principles. 

Note that higher-order (non-circular) extrapolation arcs $\curv c_{pq}$ can be constructed to fit the flow orientation estimates
at both ends exactly, implying an exactly confluent graph. 
However, some non-trivial physiological constraints have to be imposed on the smoothness/curvature of such (non-circular) confluent arcs
which should result in very long curves in cases like \figref{fig:directed weights}(c). Thus, the confluence constraint 
will manifest itself similarly to the second line in \eqref{eq:assymetric weight}. 

\begin{figure*}[t]
    \centering
    {
        \setlength{\fboxsep}{0pt}
        \setlength{\fboxrule}{0.5pt}
        \newcommand{\mygraph}[3]{
            \begin{tikzpicture}[]
              \node {\includegraphics[height=0.2\linewidth,width=0.22\linewidth,trim=145mm 70mm 170mm 20mm, clip=true]{#1}};
              #3
              \node[yshift=0.08\linewidth,xshift=0.11\linewidth,left] {\color{white} #2};
            \end{tikzpicture}
        }
        \newcommand{\colorbar}{
            \node [rectangle, bottom color=black, top color=yellow, anchor=north, minimum width=5mm, minimum height=15mm, xshift=-0.08\linewidth] (box) {};
        }
        \setlength{\tabcolsep}{0pt}
        \begin{tabular}{cc|cc}
            \multicolumn{2}{c}{\scriptsize vessel tree reconstruction using} & 
            \multicolumn{2}{c}{\scriptsize vessel tree reconstructions using} \\
            \multicolumn{2}{c}{{\bf undirected} {\em Geodesic Tubular Graph} (standard)} & 
            \multicolumn{2}{c}{{\bf directed} {\em Confluent Tubular Graph} (our)} \\[1ex]
            \mygraph{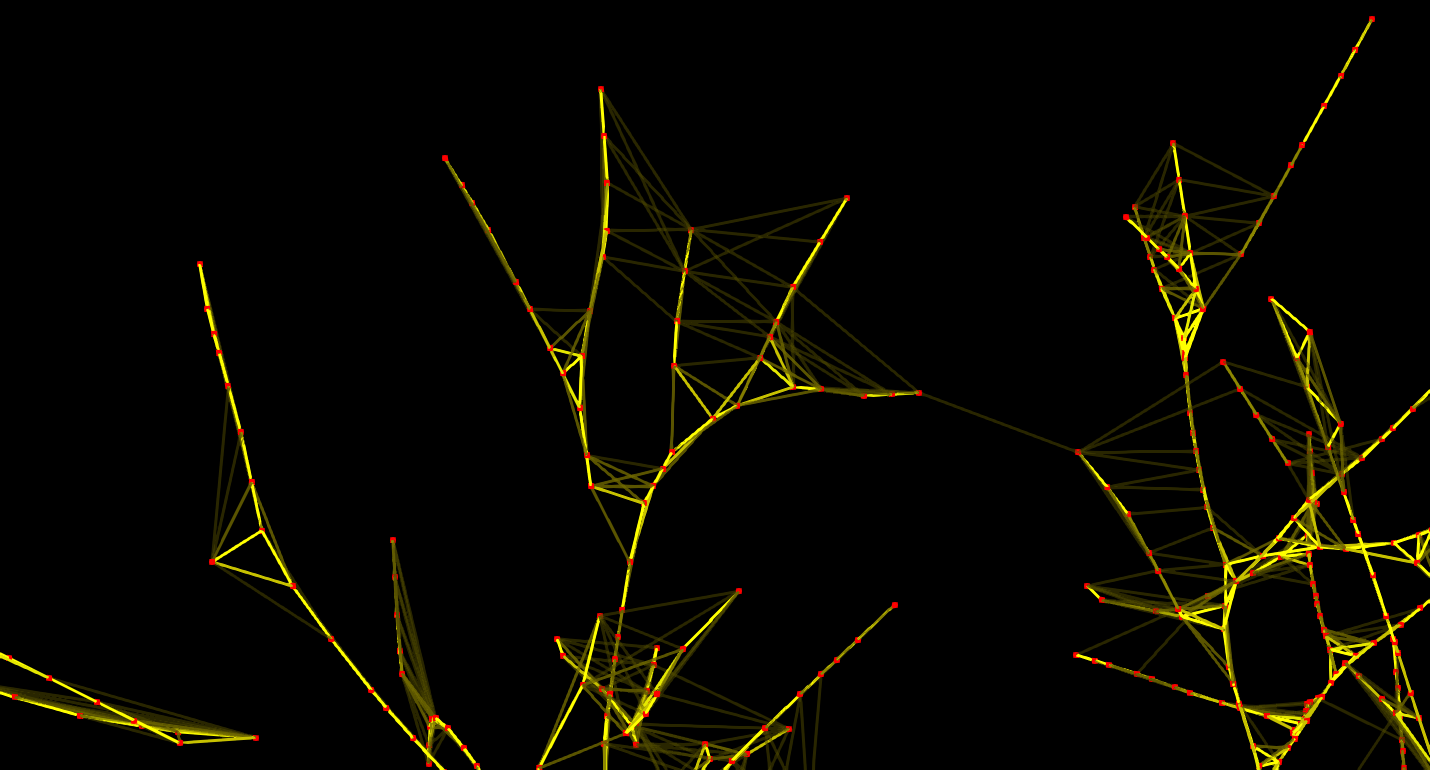}{\scriptsize \parbox{17ex}{\centering geodesic arcs}}{} & \mygraph{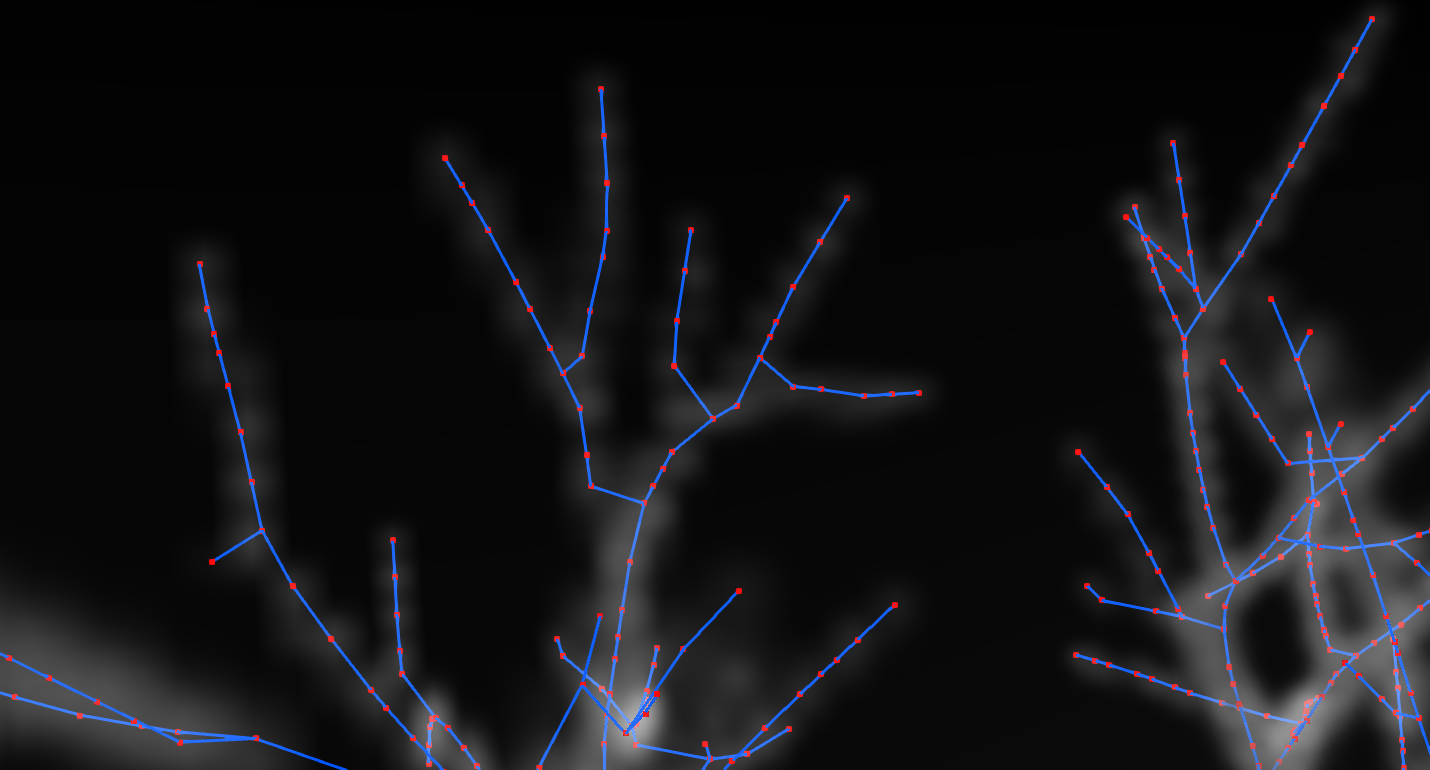}{\scriptsize GridMST}{} \hspace{1ex} & \hspace{1ex}
            \mygraph{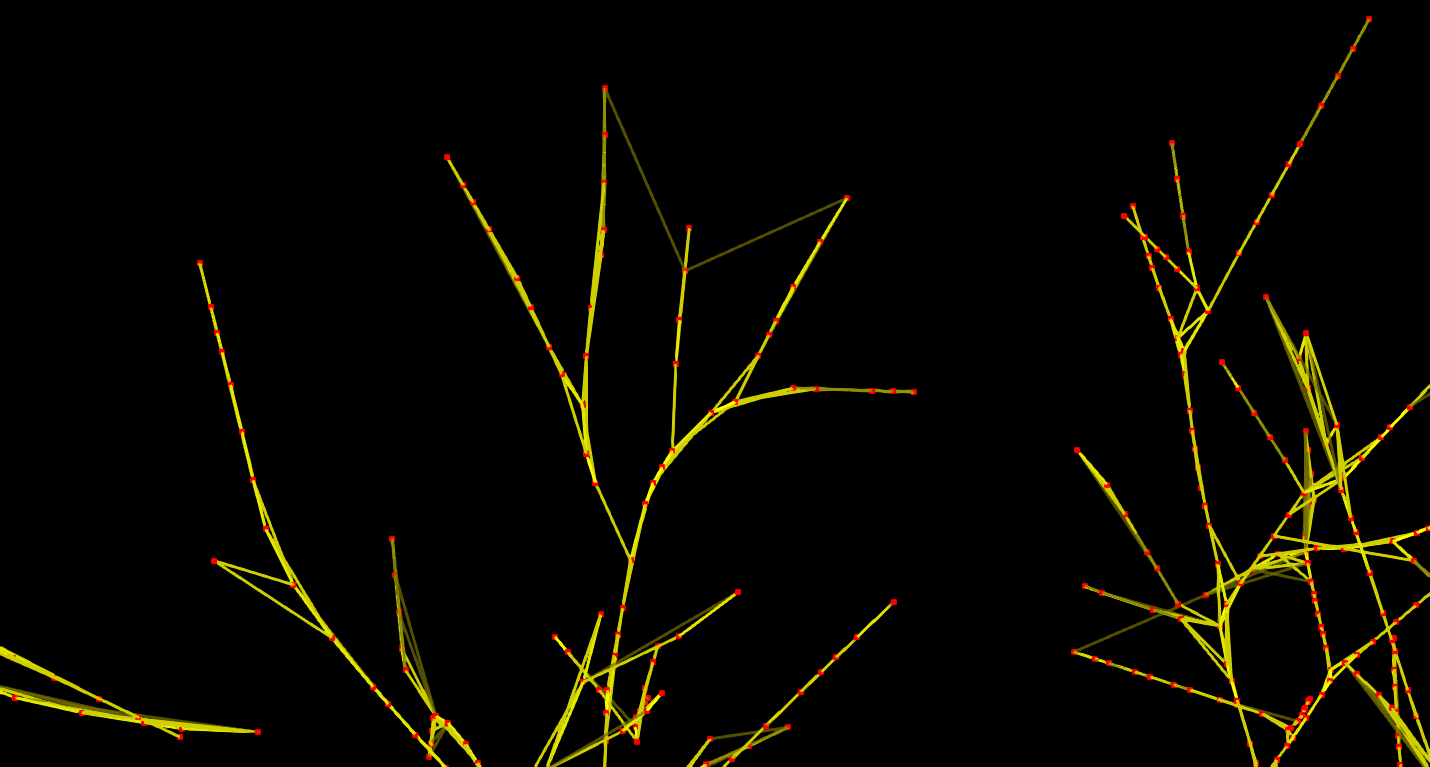}{\scriptsize \parbox{17ex}{\centering confluent arcs}}{} & \mygraph{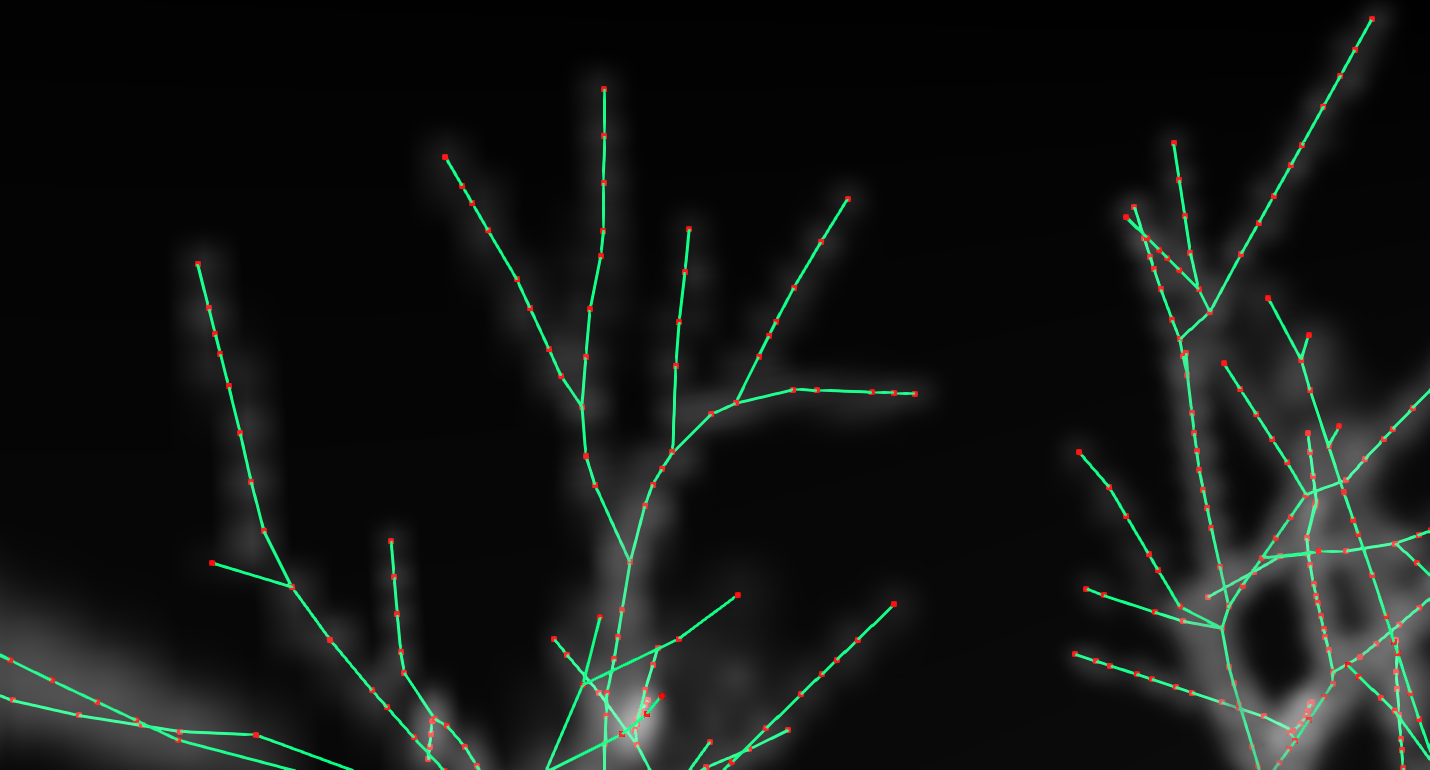}{\scriptsize MArb}{} \\ 
            \multicolumn{2}{c}{\scriptsize (a) MST (blue) for {\em Geodesic Tubular Graph} (yellow) \cite{jomier2005automatic,turetken:Neuroinform2011,Divergence:cvpr19}} &
            \multicolumn{2}{c}{\scriptsize (b) Min Arborescence (green) for {\em Confluent Tubular Graph} (yellow)} \\
        \end{tabular}
    }
    \caption{Typical tree reconstruction examples for standard geodesic (a) and our {\em confluent} (b) tubular graphs. Arc weights are represented via thickness (yellow). The data is a (representative) crop with near-capillary vessels at a periphery of large volumes, \eg \figref{fig:teaserImages}, \ref{fig:real data}.
    Sub-voxel vessels have bifurcations sparsely sampled by tubular graph nodes, as in \figref{fig:tubular graph illustration}. MST (blue) on geodesic graph ``short-cuts" most bifurcations. {\em Minimum arborescence} (green) on a directed tubular graph with {\em confluent} arcs (b), see \secref{sec:tubular_dir}, reconstructs flow-consistent bifurcations. \label{fig:gridarb}}
\end{figure*}
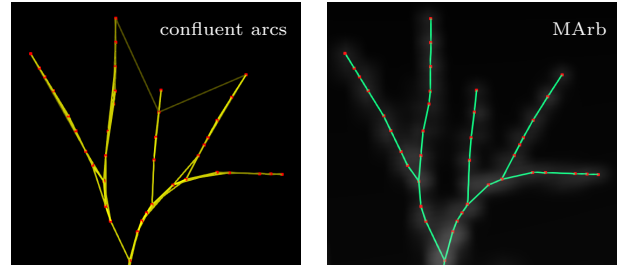

\subsection{Confluence and co-circularity} \label{sec:cf-cs}

Specifically for oriented \underline{circular} arcs, confluence implies several interesting properties and can be juxtaposed with 
the standard concept of {\em co-circularity} \cite{zucker:89}. Assume some circular flow-extrapolating arc $\curv c_{pq}$ and 
its reverse $\curv c_{qp}$ defined by two oriented tangents $\bar l_p,\bar l_q$, see \secref{sec:tubular_dir} and \figref{fig:directed weights}(a,b). 

\begin{property} 
\label{property:angle_identity}
The angle between $\curv c_{pq}^{\,1}$ and $\bar l_q$ is equal to the angle between $\curv c_{qp}^{\,1}$ and $\bar l_p$. That is,
\begin{equation} \label{eq:angle_identity}
     \angle ( \curv c_{pq}^{\,1}, \curv c_{qp}^{\,0} ) \; \equiv \; \angle (\curv c_{qp}^{\,1}, \curv c_{pq}^{\,0} ).
\end{equation}
\end{property}
While not immediately obvious, particularly in 3D, this property is not difficult to prove, see Appendix~\ref{app:proof}.
Identity \eqref{eq:angle_identity} implies the following.
\begin{theorem} \label{th:symmetry}
For circular flow extrapolating arcs, $\breve c_{pq}$ is confluent iff the reverse 
 arc $\breve c_{qp}$ is confluent.
\end{theorem}
This theorem shows that confluence of $\breve c_{pq}$ and $\breve c_{qp}$ in \figref{fig:directed weights}(a,b) is not a coincidence. 
However, in general, such symmetry does not hold for non-circular flow-extrapolating arcs (higher order polynomial curves, etc). 
Also, Theorem \ref{th:symmetry} does not imply ``undirectedness" of our confluent tubular graph construction using simple 
circular arcs. As follows from \eqref{eq:assymetric weight}, $w_{pq} \neq w_{qp}$ 
since the reverse arcs $\breve c_{pq}$ and $\breve c_{qp}$ have different lengths regardless of confluence, 
see \figref{fig:directed weights}(a,b).

Interestingly, our confluence constraint in case of {\em circular} oriented arcs can be related to an ``oriented'' generalization of 
{\em co-circularity} that was originally defined in \cite{zucker:89} for 2D curves. 
In $\mathbb R^n$ co-circularity constraint can be defined for two {\em unoriented} tangent lines $l_p$ and
$l_q$ at points $p$ and $q$ in a way similar to our definition of confluence for $\curv c_{pq}$ that is based on 
{\em oriented} tangents $\bar l_p$ and $\bar l_q$. Assume {\em unoriented} circle $c_{pq}$ uniquely defined in $\mathbb R^n$
by a pair of points $p,q$ and tangent $l_p$ at the first point. If we use $c^{x}$ to denote an unoriented unit tangent of 
circle $c$ at any given point $x$, then circle $c_{pq}$ is uniquely defined by three conditions
\begin{equation}
c_{pq} \;\;\;:\;\;\;\; p\in c_{pq},\;\;q\in c_{pq}, \;\; c_{pq}^{\,p}=l_p. 
\end{equation}
In general, circle $c_{qp}$ is different 
as it is defined by tangent $l_q$ at point $q$, that is $c_{qp}^q=l_q$. 
Then, co-circularity constraint for $l_p$ and $l_q$ can be defined as 
\begin{equation} \label{eq:co-circ}
     \angle ( c_{pq}^{\,q}, c_{qp}^{\,q} ) \;\equiv\; \angle ( c_{qp}^{\,p}, c_{pq}^{\,p} )   \;\leq\; \epsilon
\end{equation}
where $\angle(\cdot,\cdot)$ is the angle between two lines  in contrast to the angle between vectors in the similar identity \eqref{eq:angle_identity}.

The difference between confluence for $\bar l_p$, $\bar l_q$ and co-circularity for $l_p$, $l_q$ can be illustrated by the
examples in \figref{fig:directed weights}. Note that unoriented versions of $\bar l_p$, $\bar l_q$ are identical in
all three examples (a,b,c) as they do not depend of the flip of orientation in (c). Thus, they are equally co-circular 
in (a,b,c). At the same time, oriented tangents are confluent in (a,b) while flipping orientation for $\bar l_p$ results in
non-confluence in (c). The properties discussed above imply that confluence can be seen as oriented generalization of co-circularity 
\cite{zucker:89}. 
\begin{property}
Confluence of oriented circular arcs $\breve c_{pq}$ or $\breve c_{qp}$, which are defined by oriented tangents $\bar l_p, \bar l_q$, 
implies co-circularity of the corresponding unoriented tangents $l_p, l_q$, 
but not the other way around. 
\end{property}

Note that co-circularity constraint for {\em unoriented} circular arcs along a path on a tubular graph can enforce $\mathcal G^1$-smoothness 
within a single vessel branch. But, unoriented co-circularity enforces smoothness indiscriminately in all directions from a bifurcation point without
resolving conflicts between multiple branches. This leads to artifacts observed on geodesic tubular graphs, see \figref{fig:gridarb}(a).
In contrast, the confluence constraint discriminates orientations of branches when enforcing smoothness at bifurcations, 
see \figref{fig:gridarb}(b).

\subsection{Confluent tree reconstruction algorithm} \label{sec:alg}

\begin{algorithm}
    \begin{algorithmic}[1]
        \REQUIRE Raw volumetric data and root location
        \STATE Estimate a set of centerline points $V$ and directed flow estimates $\bar L=\{\bar l_p|p\in V\}$.
        \STATE Build a set of oriented arcs $A \subseteq V\times V$, \secref{sec:tubular_dir}
        \STATE Build a \textit{confluent tubular graph} $G$ by computing 
                weights $w_{pq}$ for $(p,q)\in A$ using \eqref{eq:assymetric weight}.
        \STATE Return the \textit{Minimum Arborescence} of $G$. 
    \end{algorithmic}
    \caption{Confluent Tree Reconstruction}
    \label{alg:main}
\end{algorithm}

Our Confluent Tree Reconstruction Algorithm \ref{alg:main} is discussed below. It inputs raw volumetric data with a marked root of the tree. The algorithm has four steps. First, it runs a subroutine that estimates a set of points on the tree centerline $V$ and 
oriented flow pattern at these points $\bar L$. 
We use a standard vector field estimation method \cite{Divergence:cvpr19} based on non-negative
divergence constraint and regularization over the voxel-grid neighborhood. Since $80\%$ of our large
vasculature volumes are near-capillary vessels, the weak sub-voxel signal often results in missing data points 
and grid-based regularization fails to produce consistent flow orientations, particularly at the tree periphery. 
We modified \cite{Divergence:cvpr19} by (anisotropically) enlarging their regularization neighborhood, see Appendix~\ref{app: ANN}, 
improving the quality of flow estimates $\bar L$ that helps to reconstruct confluent vessel trees. 

Second, we build a set of oriented arcs between the points in $V$ that correspond to directed edges on our tubular graph $G$. Our confluence constraint works well even with a complete graph $A=V\times V$. But, for efficiency, we restrict the neighborhood 
to $K$ nearest neighbors (KNN). The running time is $\mathcal O(K|V|\log|V|)$ with $k$-d trees.

The last two steps compute a directed weight $w_{pq}$ for all arcs $(p,q) \in A$ as described in \secref{sec:tubular_dir}, and invoke a standard minimum arborescence algorithm that has complexity $\mathcal O(|A| + |V|\log|V|)$ \cite{gabow1986efficient}.
In practice, the overall running time of our method for vessel tree reconstruction is dominated by the centerline localization and flow pattern estimation in the first step.

\section{Experimental results} \label{sec:experiments}


\label{sec:impl details}


We use two baselines which we call NMS-MST and GridMST. NMS-MST uses the Frangi method \cite{frangi1998multiscale} along with Non-Maximum Suppression (NMS) to obtain the centerline points and local unoriented tangent estimates. On top of these, NMS-MST uses the KNN ($K=500$) graph of the centerline points to build the MST. Note that the KNN graph is symmetric such that a pair of nodes have an arc as long as one is a neighbor of the other. Here, the undirected edge weight is computed by the sum of two shorter arc lengths. 
GridMST uses \cite{Divergence:cvpr19} for estimating the centerline points and flow direction. Then, it also uses the KNN graph of the centerline points to build the MST. 

GridArb uses the set of centerline points and flow estimates produced by \cite{Divergence:cvpr19}, but it uses the confluent tubular graph to build the minimum arborescence (discussed in \secref{sec:alg}). MArb exploits a modified version of \cite{Divergence:cvpr19} (see Appendix~\ref{app: ANN}) and also uses the confluent tubular graph to build the minimum arborescence. We set $\varepsilon=\frac{\pi}{2}$ in \eqref{eq:assymetric weight} for all our experiments.
\vspace{1pt}

\noindent
\begin{tabular}{m{1.35cm}|m{2.5cm}|m{1.35cm}|m{1.5cm}}
    \bf Method & \bf Flow \mbox{estimates} & \bf Graph Weights & \bf Tree Extraction\\ 
    \hline
    \mbox{NMS\!-\!MST} & Frangi~\etal\,\cite{frangi1998multiscale} &  &  \multirow{2}{1.7cm}{MST} \\
    \cline{1-2}
    GridMST & Zhang~\etal\,\cite{Divergence:cvpr19} & \multirow{-2}{1.5cm}{standard geodesic} &  \\
    \hline
    GridArb & Zhang~\etal\,\cite{Divergence:cvpr19} & \multirow{3}{1.7cm}{our \mbox{confluent} \eqref{eq:assymetric weight}} & \multirow{3}{1.7cm}{minimum arbores-cence} \\
    \cline{1-2}
    MArb & Modified~\cite{Divergence:cvpr19},~see suppl materials & & \\
\end{tabular}

\subsection{Validation Measures} \label{sec:val}

Many validation measures rely on matching between the ground truth and the predicted tree. Matching algorithms could be separated into several groups. First, match the nodes of the trees independently based on a distance measure, \eg \cite{Divergence:cvpr19,turetken2013detecting,tyrrell:07}, partial (local) sub-tree matching \cite{gillette2011diadem}, or global tree matching approaches \cite{klein1998edit,zhang1989simple,chesakov2015vascular}. We base our evaluation approach on the first group of methods due to their efficiency and the size of our problem.

\textbf{Centerline reconstruction quality.}
Our reconstructed tree is ideally the centerline of the vasculature. We compute the recall and fall-out statistics of the centerline points to evaluate the reconstruction quality. To obtain the \emph{centerline receiver operating characteristic (ROC) curve}, we generate a sequence of \emph{recall}/\emph{fall-out} points by varying the \emph{detection threshold} parameter for the low-level vessel filter of Frangi \etal \cite{frangi1998multiscale}.

Similarly to \cite{Divergence:cvpr19}, a specific point on the ground truth centerline is considered detected correctly (recall) iff it is located within $\max(r, \zeta)$ distance of a reconstructed tree where $r$ is the radius of the corresponding ground truth vessel segment and $\zeta = \frac{\sqrt{2}}{2}$ voxel-size. A point on the reconstructed tree that is farther away than this distance is considered incorrectly detected (fall-out).
Before computing the ROC curve we re-sample uniformly both the ground truth and reconstructed trees.

\textbf{Bifurcation reconstruction quality.}
We introduce two separate metrics. First, we compute the \emph{ROC} curve for only bifurcation points to assess the quality of detection. 
Second, we measure the median angular error at the reconstructed bifurcations to assess the accuracy, where we match \emph{all} ground truth bifurcations to closet branching points on the detected tree regardless of their proximity and use the median rather than the average for greater stability. The difference between our angular error measure and that in \cite{Divergence:cvpr19} is discussed in Appendix~\ref{app:anlge error}.



\begin{figure}[t]
    \centering
    \small
    \begin{tabular}{m{4ex}m{0.4\linewidth}m{0.4\linewidth}}
    \rotatebox{90}{noise \emph{std} 10} &
    \includegraphics[width=\linewidth]{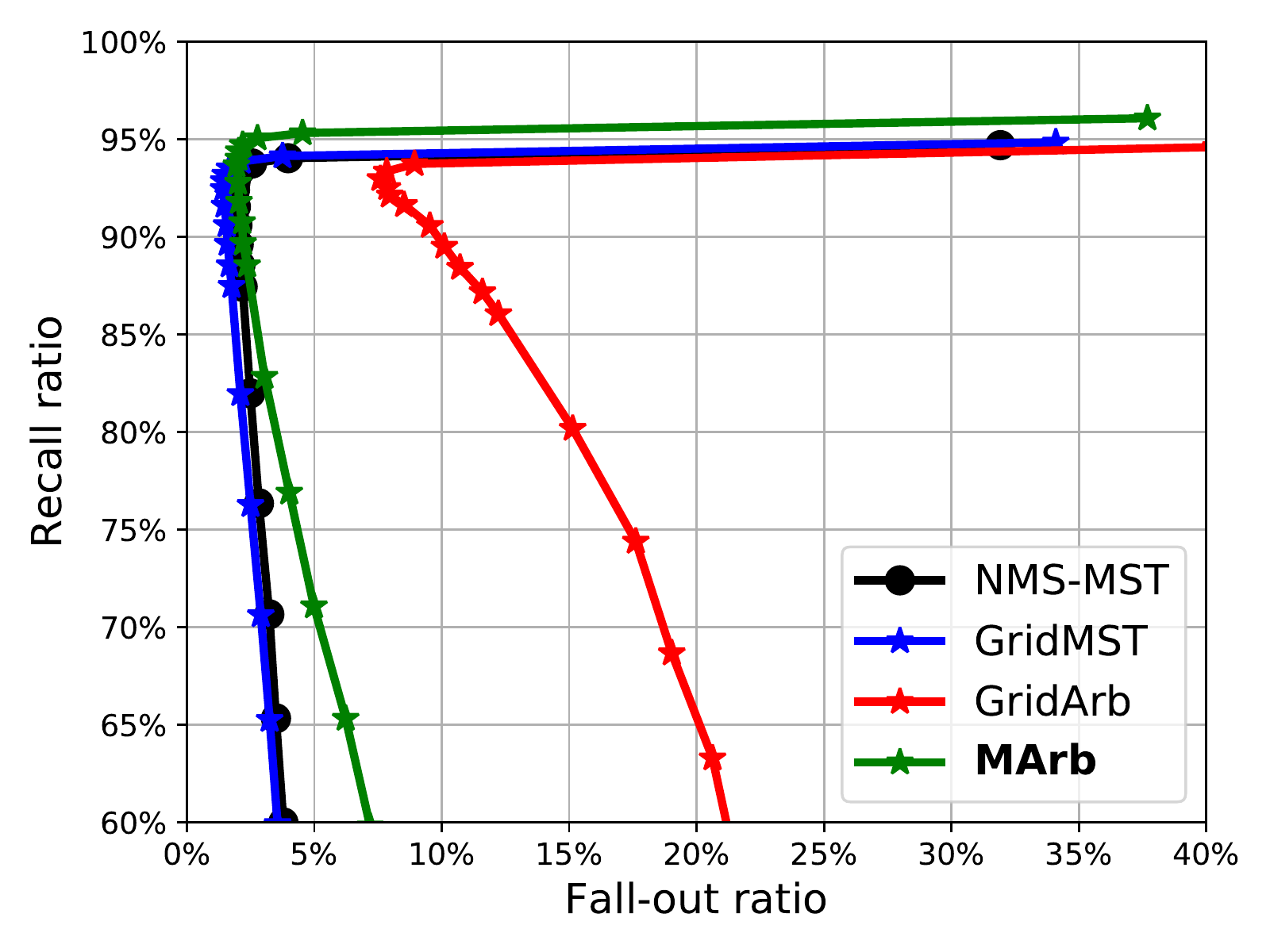} &
    \includegraphics[width=\linewidth]{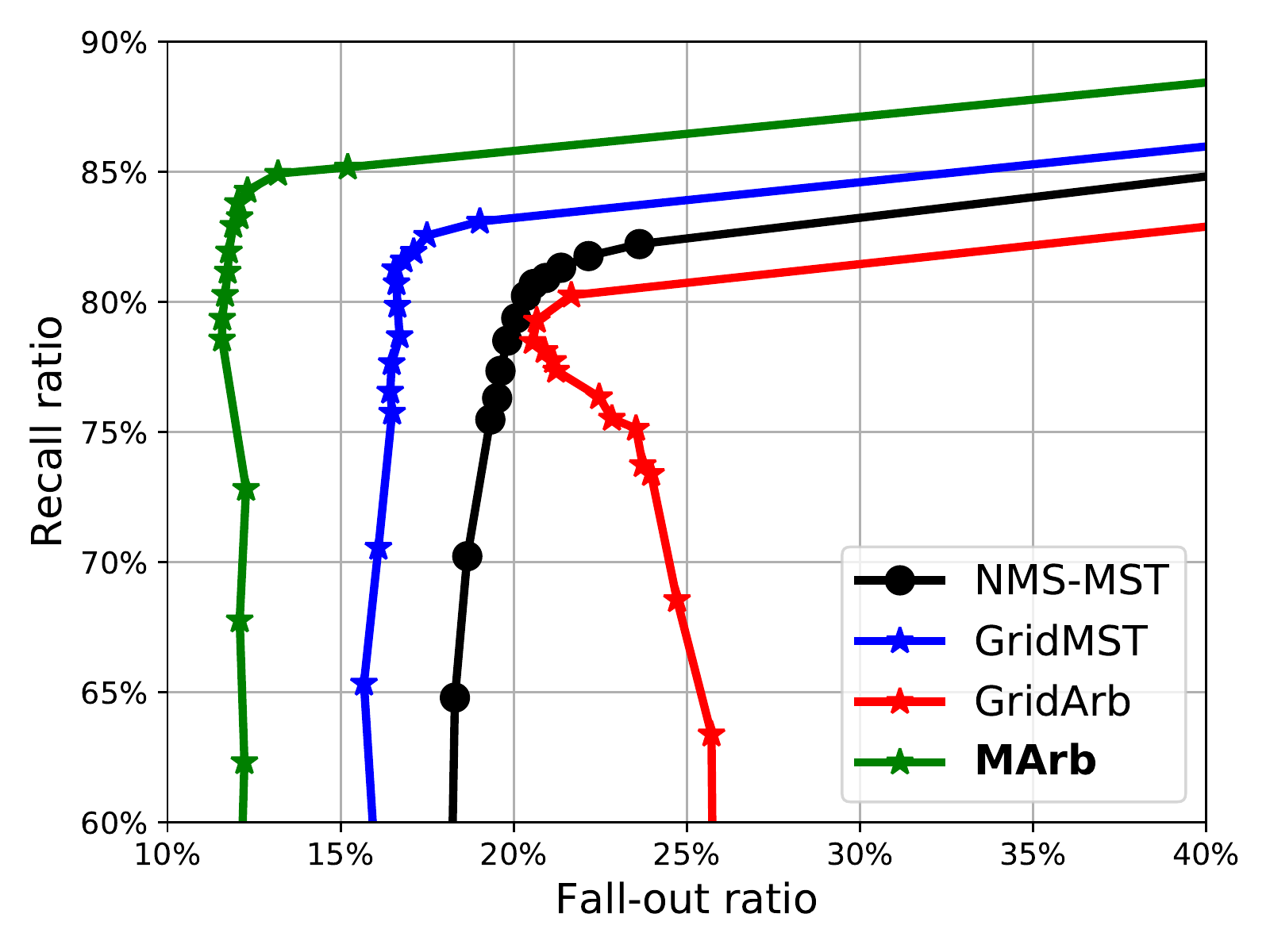}\\
    \rotatebox{90}{noise \emph{std} 15} &
    \includegraphics[width=\linewidth]{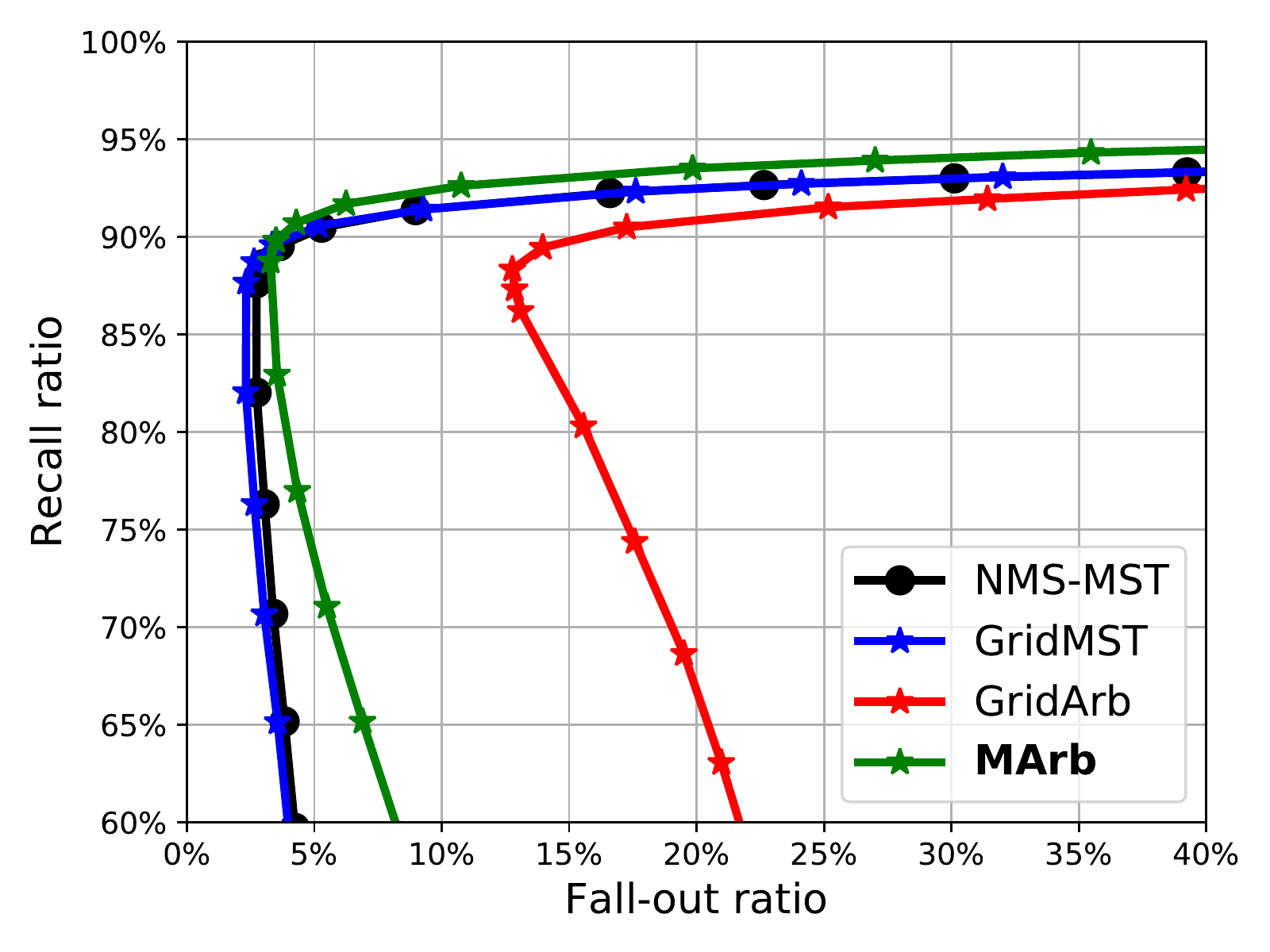} &
    \includegraphics[width=\linewidth]{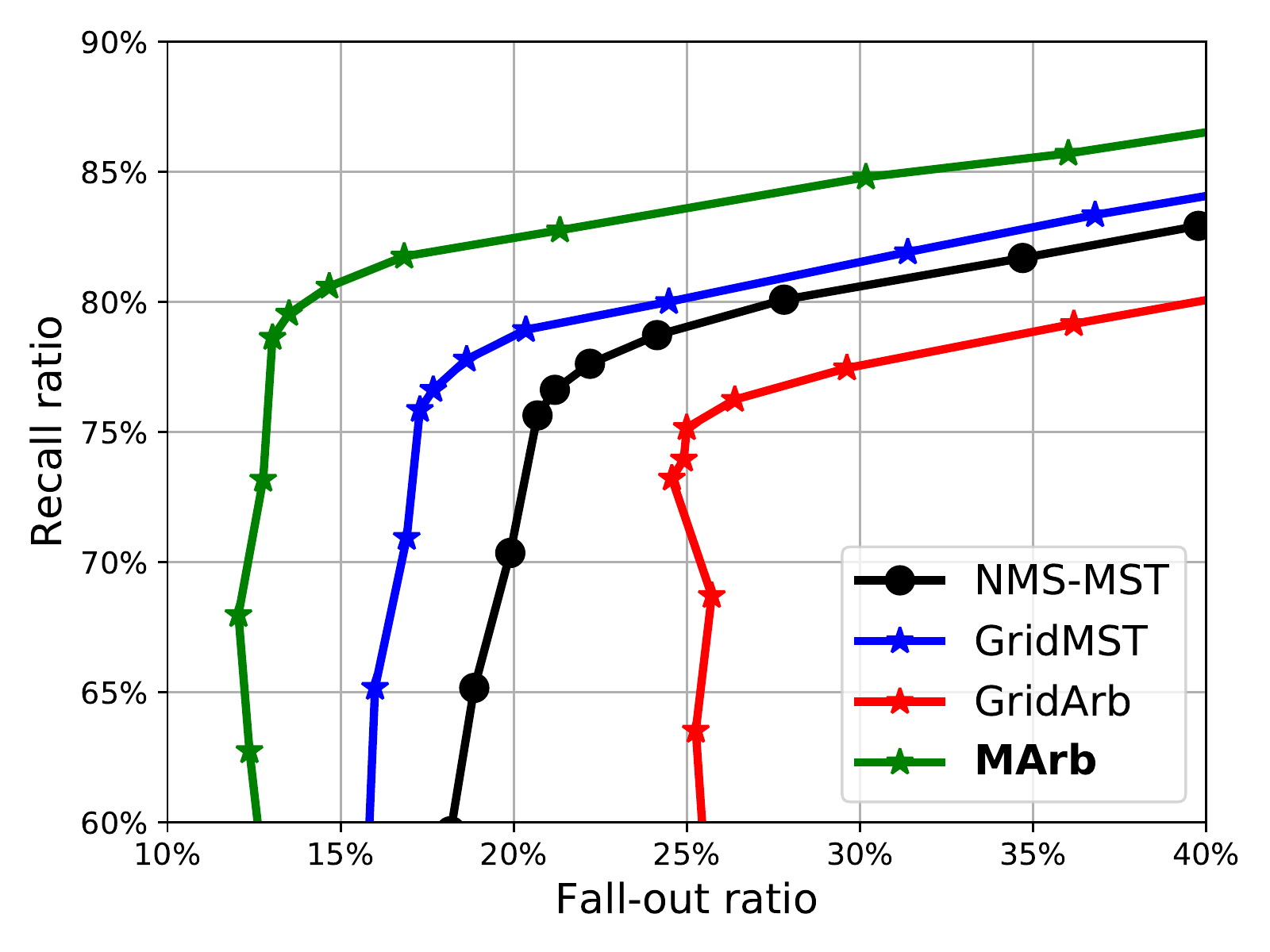}\\
    & \centering (a) centerline detection & \centering (b) bifurcation detection
    \end{tabular}
    \vspace{-2mm}
    \caption{Quantitative comparison. Our methods are denoted by MArb and GridArb. GridMST is the best result from \cite{Divergence:cvpr19}.}
    \label{fig:roc curves}
\end{figure}
\begin{figure}[t]
    \centering
    \small
    \begin{tabular}{cc}
    noise \textit{std} 10 & noise \textit{std} 15 \\[-0.5ex]
    \includegraphics[width=0.5\linewidth,clip=true,trim=0 0 0 3mm]{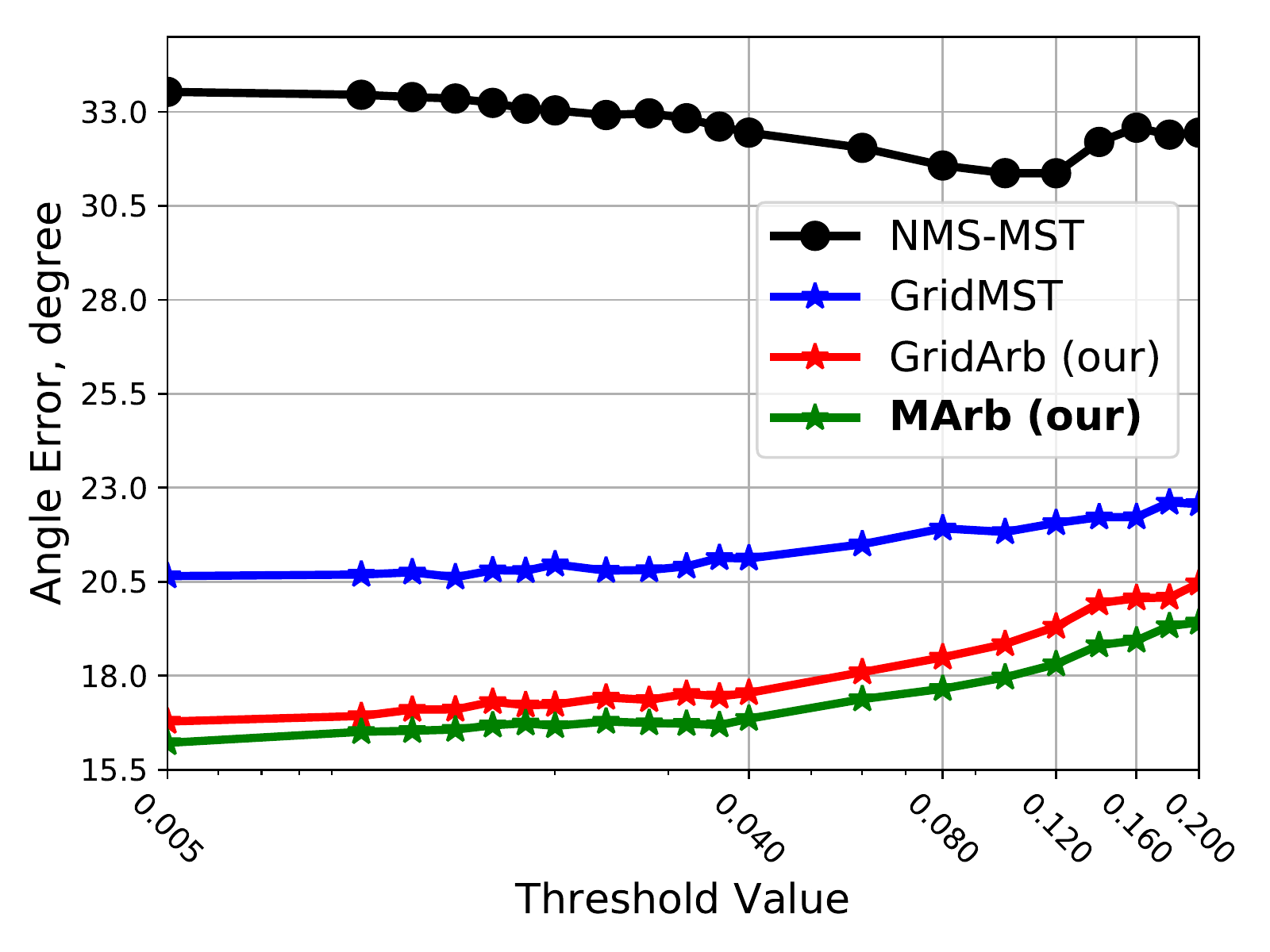} &
    \includegraphics[width=0.5\linewidth,clip=true,trim=0 0 0 3mm]{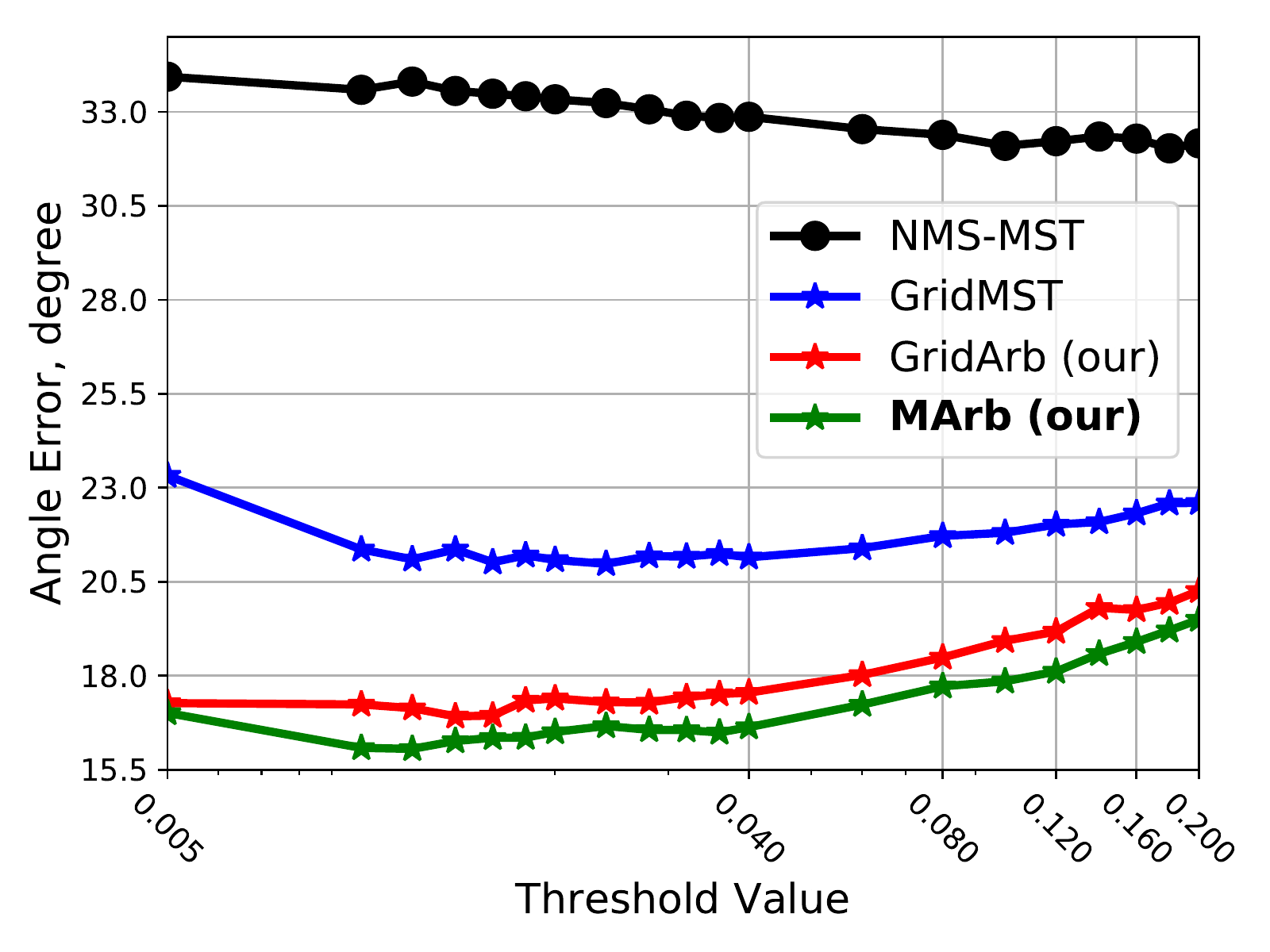}
    \end{tabular}
    \vspace{-5mm}
    \caption{Median branching angular error for our methods (GridArb and MArb from \secref{sec:method}) outperform all competitors (NMS-MST and GridMST \cite{Divergence:cvpr19})}    \label{fig:angular errors}
\end{figure}

\subsection{Synthetic Data with Ground Truth}\label{sec:synthetic data}
One of the major challenges in large-scale vessel tree reconstruction is the lack of ground truth. That complicates many interactive and supervised learning methods and makes evaluation hard. Zhang \etal \cite{Divergence:cvpr19} generated and published a dataset with ground truth using \cite{hamarneh2010vascusynth}. We used our newly generated 15 volumes (see Appendix~\ref{app:synthetic data}) $100\!\times\!100\!\times\!100$ with intensities between $0$ and $512$. Our new dataset has a larger variance of bifurcation angles. The voxel size is $0.046$ mm. We add Gaussian noise with {\em std} 10 and 15. 

\figref{fig:roc curves} compares the results of our methods with two competitors. One is the method of \cite{Divergence:cvpr19}, another baseline is simple MST computed over non-maximum suppression of vessel filter output. All methods use essentially the same detection mechanism, \ie Frangi \etal filter, so the centerline extraction quality does not differ much. On the other hand, our method significantly outperforms in the quality of bifurcation detection, see \figref{fig:roc curves} (b). This result is complemented by superior angular errors in \figref{fig:angular errors}. We attribute this to the subvoxel accuracy and better reconstruction of bifurcation. A typical example is shown in \figref{fig:teaserImages}(b,c).

The GridArb performs competitively in terms of angular errors but gives the worst results in terms of centerline quality. This is due to the artifact caused by some inconsistent flow estimates near the tree periphery (see \figref{fig:gridarb} for concrete examples). In \secref{sec:alg} we argue that enlarging the regularization neighborhood helps improve the estimation of the flow orientation. 

\subsection{High Resolution Microscopy CT} \label{sec:real data}

\begin{figure}[t]
    \centering
    \newcommand{\myzooms}[1]{\includegraphics[height=0.15\linewidth,width=0.2\linewidth]{#1}}
    \begin{tabular}{cc}
        \multirow{6}{0.8\linewidth}{ 
            \def\big{\includegraphics[width=\linewidth,height=1.18\linewidth]{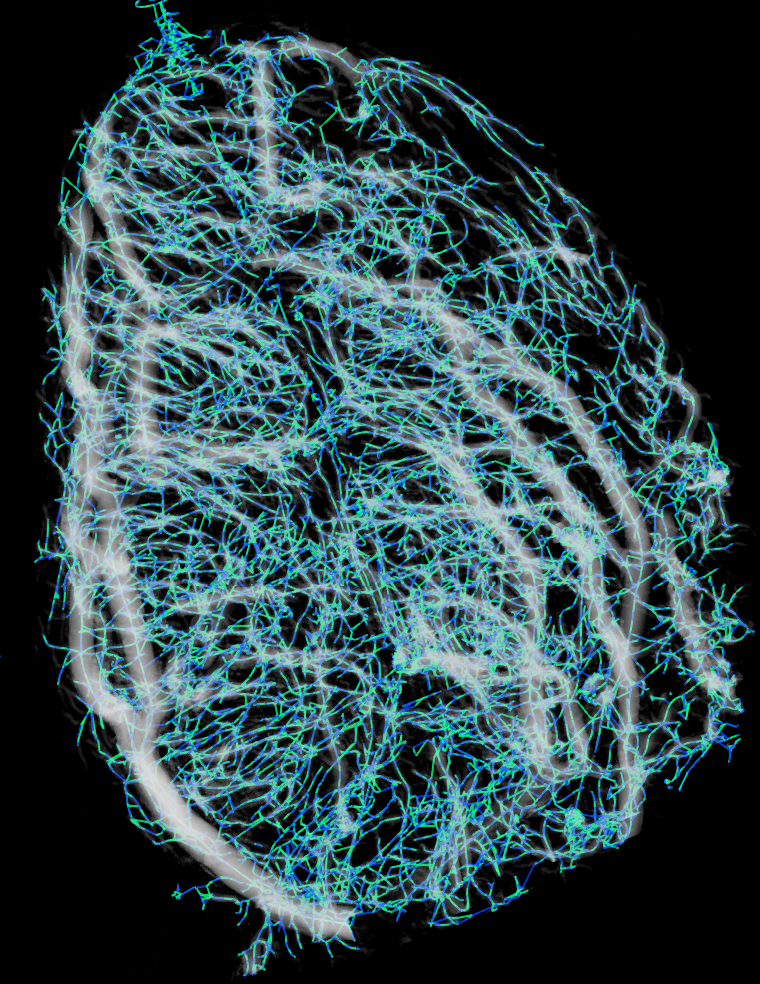}}
            \def\little{\colorbox{black}{\includegraphics[width=0.5\linewidth]{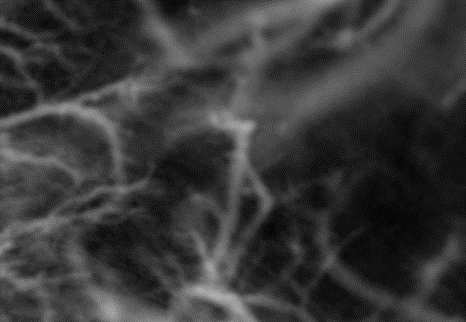}}}
            \def\stackalignment{l}
            \topinset{\little}{\big}{0pt}{0pt}
        } & \\ 
        & \myzooms{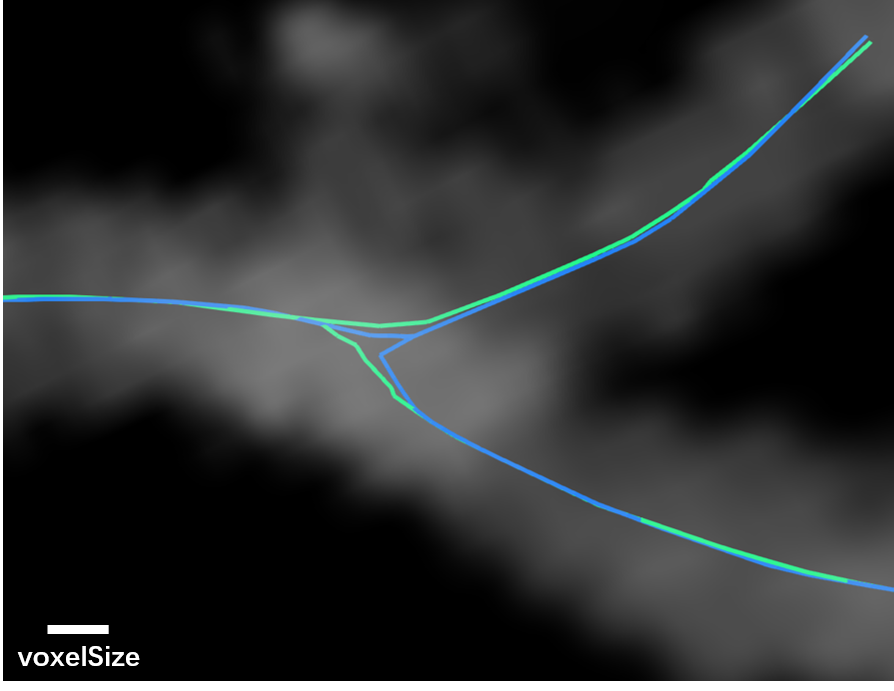} \\
        & \myzooms{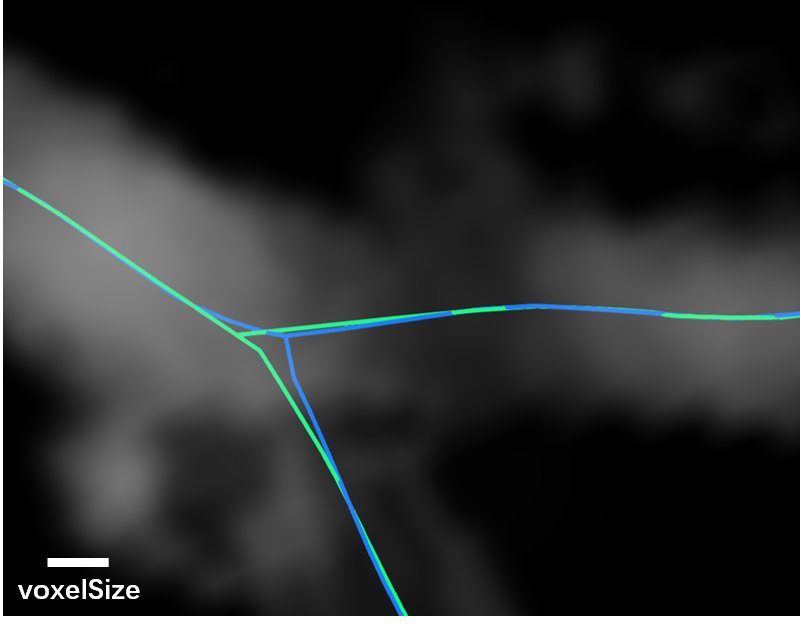} \\
        & \myzooms{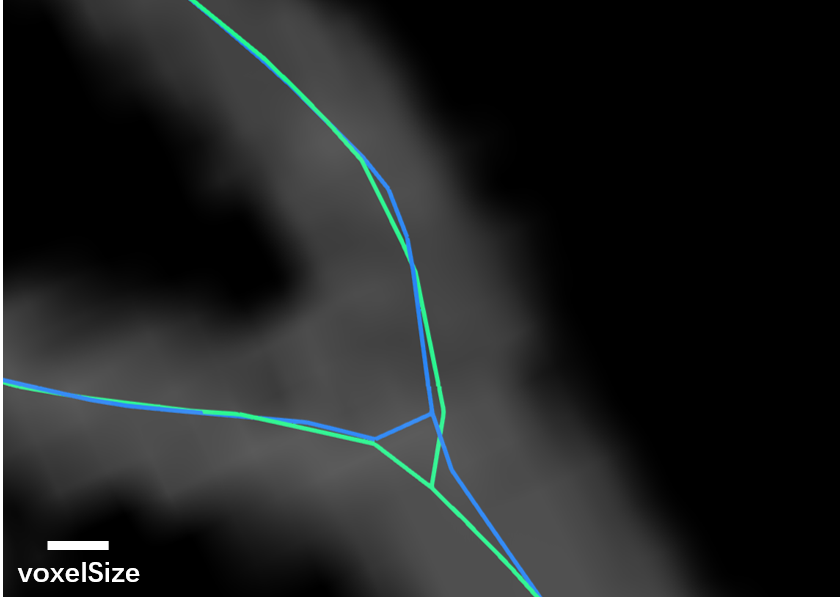} \\
        & \myzooms{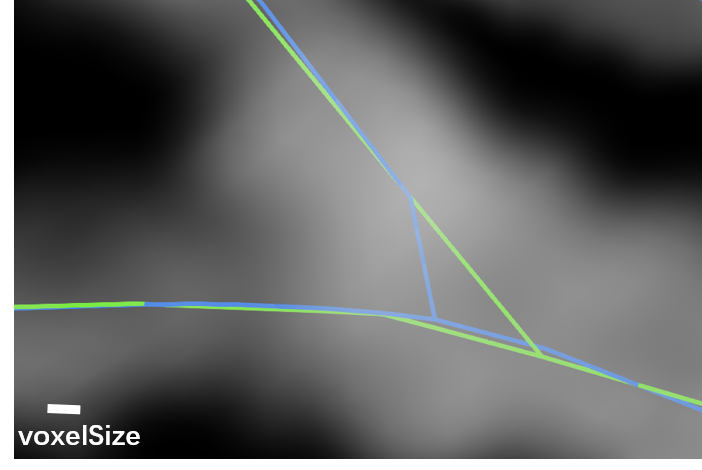} \\
        & \myzooms{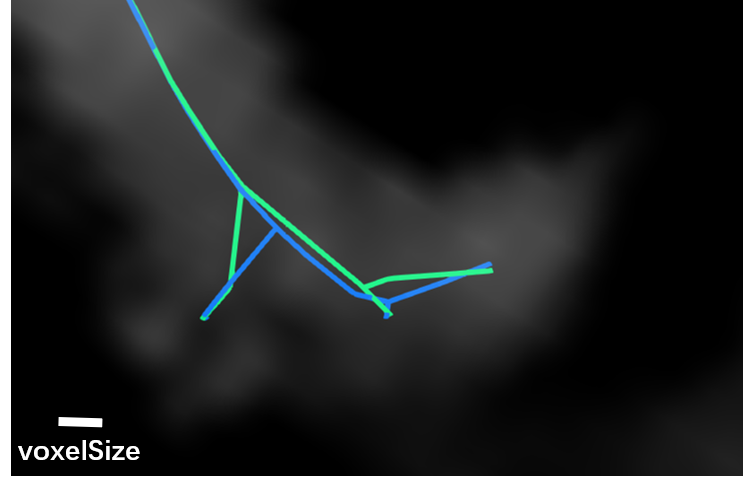} \\
        & \myzooms{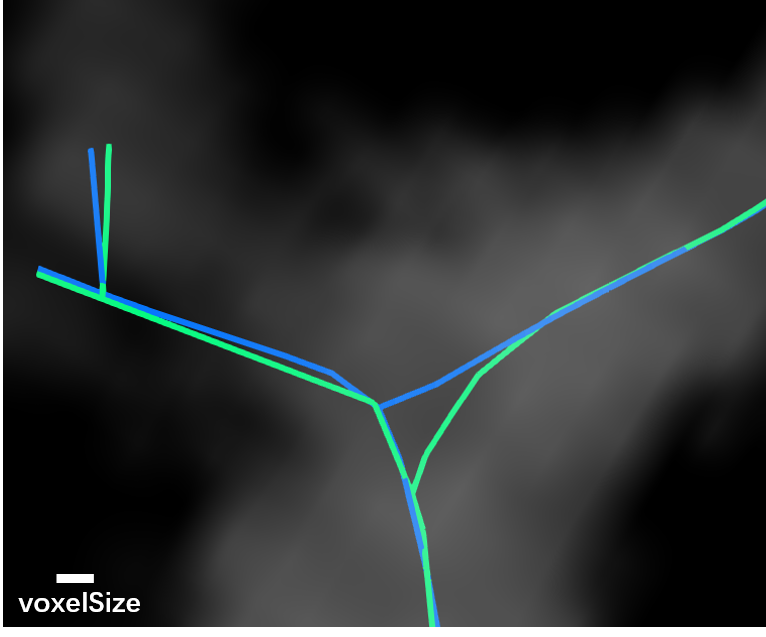} \\
        \resizebox{0.8\linewidth}{0.15\linewidth}{%
            \myzooms{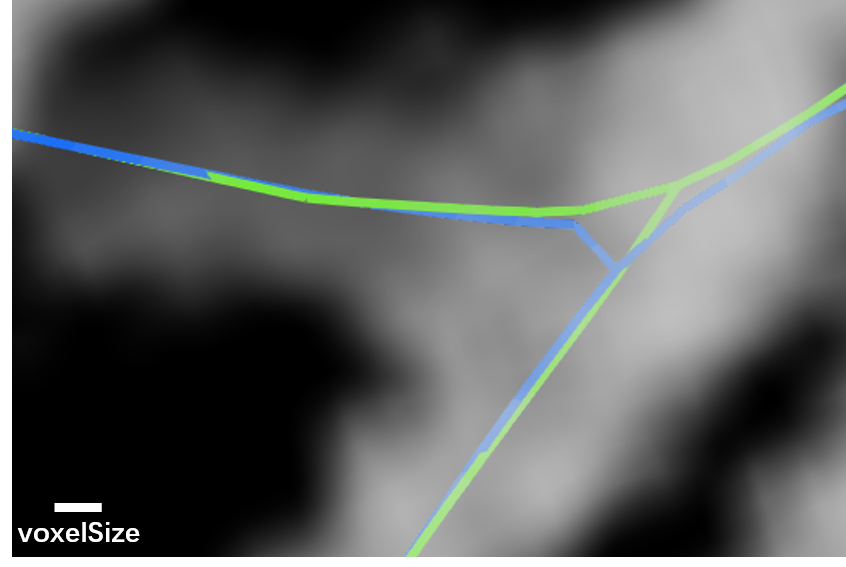} \myzooms{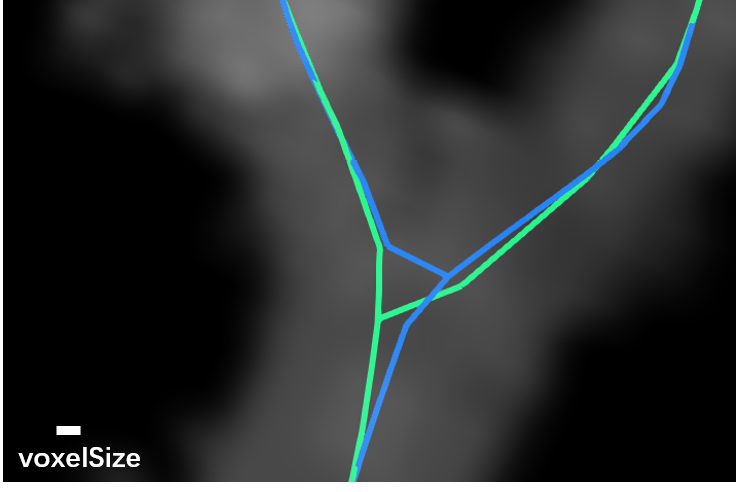} \myzooms{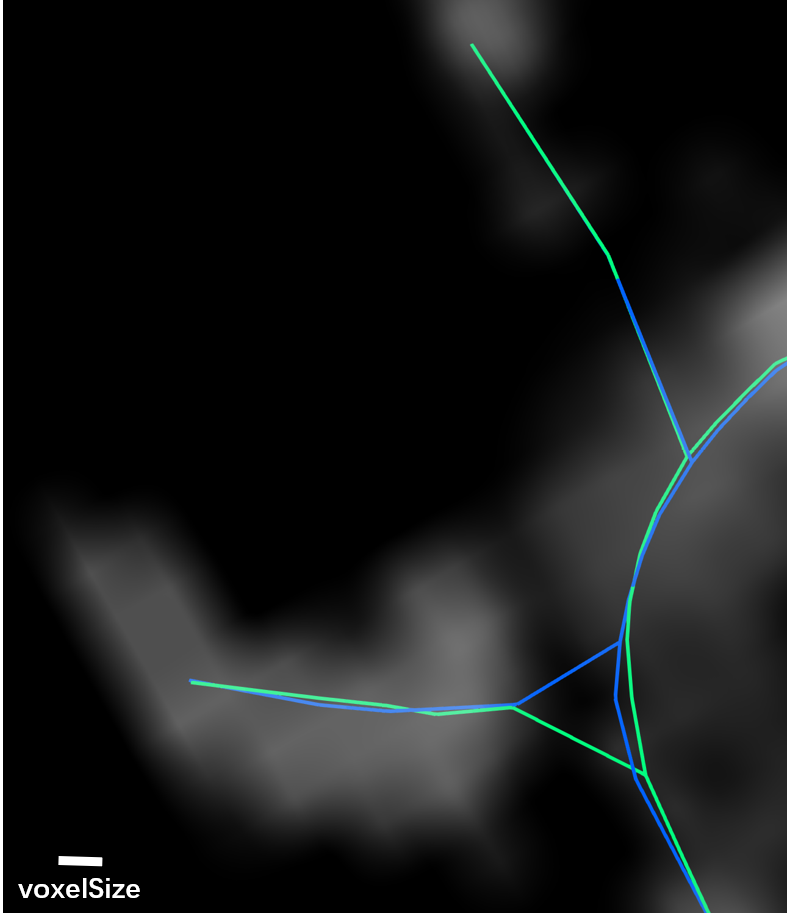} \myzooms{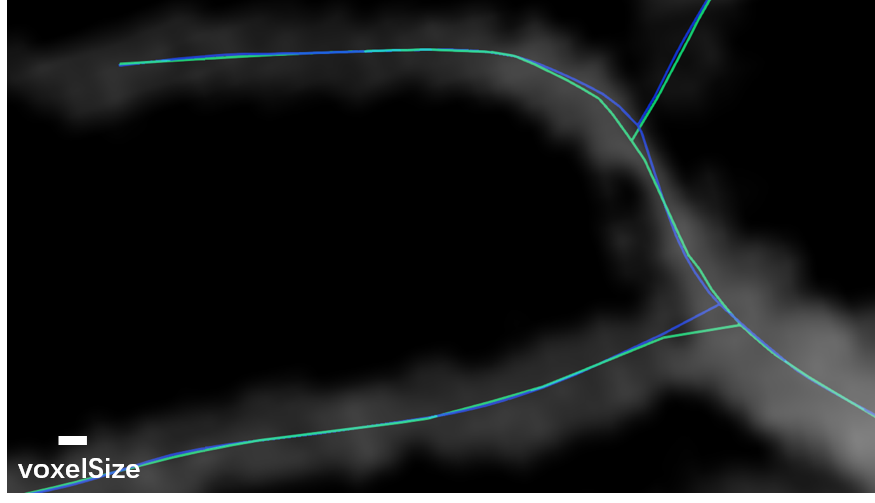}%
        } & \myzooms{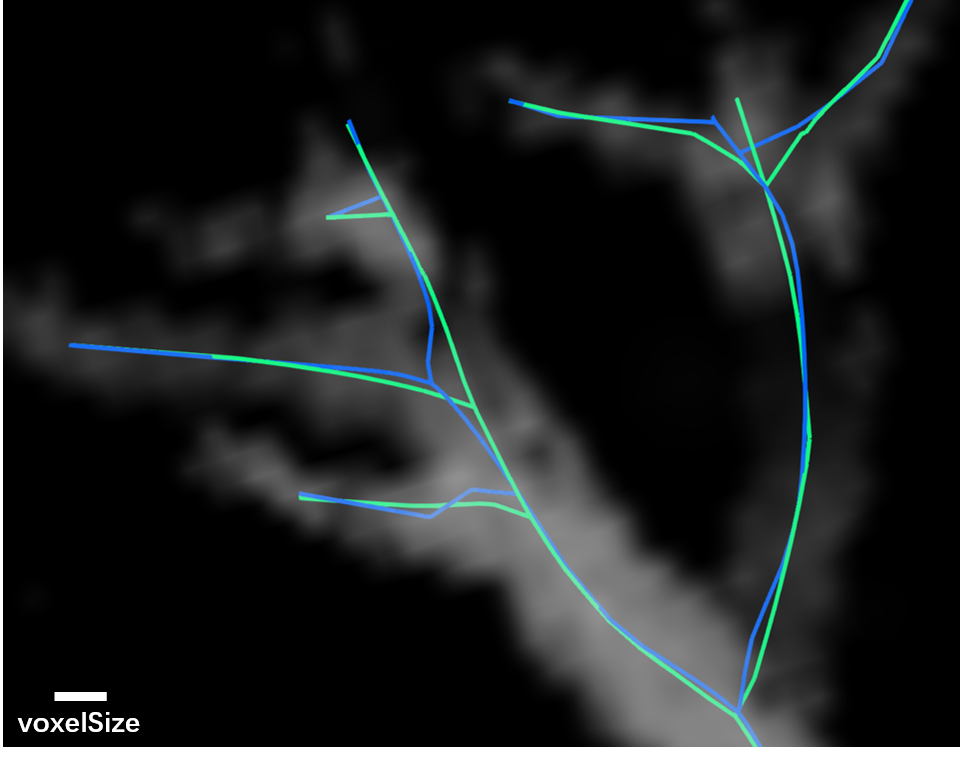}
        \end{tabular}
    \caption{Real data micro-CT volume and vessel centerline reconstruction (green) obtained by our method and (blue) by \cite{Divergence:cvpr19}. To reduce clutter we show the result of Frangi \etal filter \cite{frangi1998multiscale} instead of raw input data. The real data zoom-in is at the top left.}
    \label{fig:real data}
\end{figure}
We use challenging microscopy computer tomography volume (micro-CT) of size $585 \!\!\times\!\! 525\!\! \times\!\! 892$ voxels to qualitatively demonstrate the advantages of our approach. The data is a high-resolution image of mouse heart obtained \foreign{ex vivo} with the use of contrast. The resolution allows detecting nearly capillary level vessels, which are partially resolved (partial volume).  \figref{fig:real data} shows the whole volume and reconstruction.


\bibliographystyle{ieee_fullname}
\bibliography{ref}

\newpage
\appendix
\appendixpage
\addappheadtotoc
\section{The proof of Property~\ref{property:angle_identity}} \label{app:proof}
\begin{figure}[b]
\centering
\includegraphics[page=2,width=\linewidth,trim=70mm 35mm 50mm 35mm,clip=true]{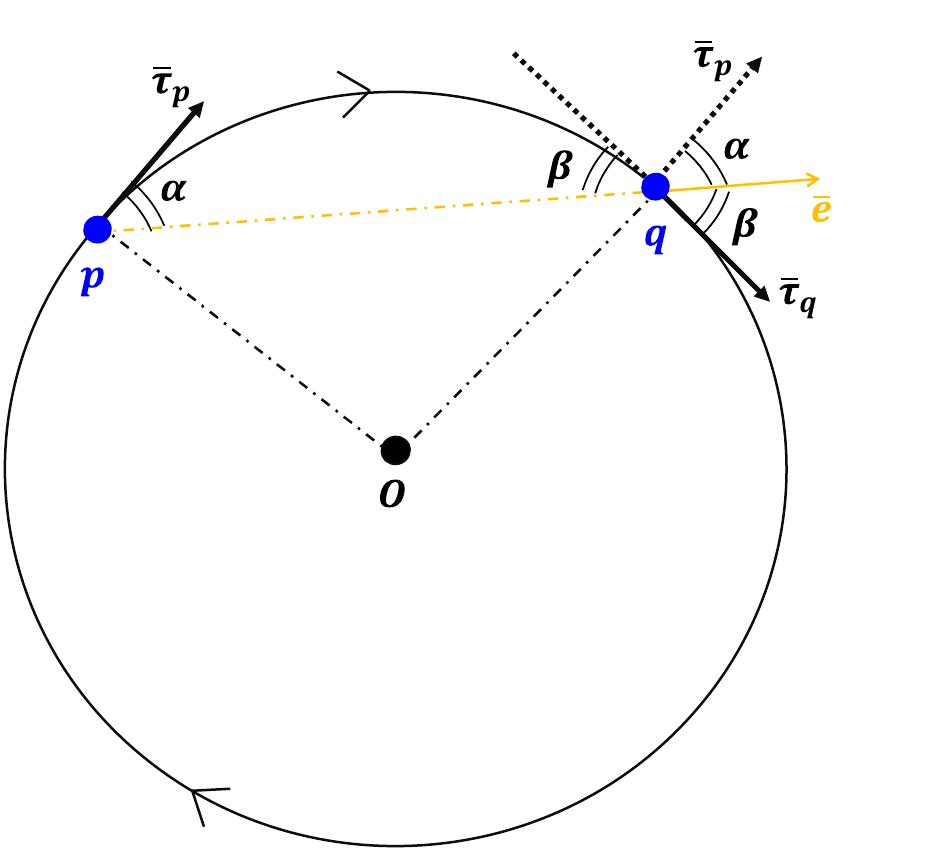}
\caption{Illustration for \eqref{eq: proof1}. $O$ is the center of the circle. $\bar e$ is a unit vector along $pq$.}
\label{fig:proof1}
\end{figure}

\begin{figure}[b]
\centering
\includegraphics[page=1,width=\linewidth,trim=70mm 35mm 50mm 35mm,clip=true]{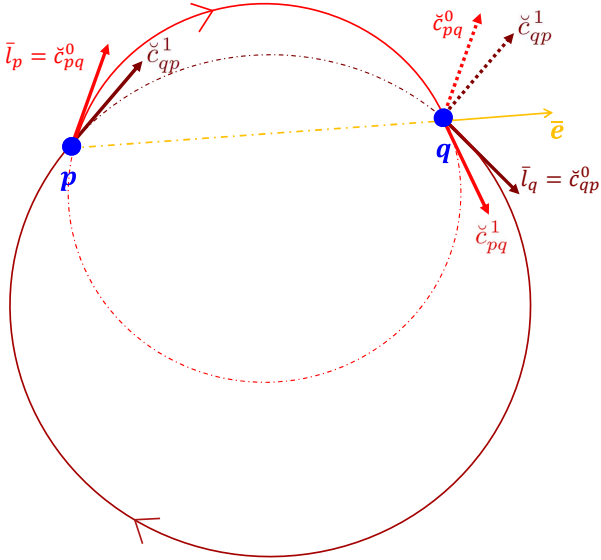}  
\caption{Two circles given by $p$,$q$, $\curv c^0_{pq}$ and $\curv c^0_{qp}$. Vectors in red are co-planar with the red circle while those in maroon are co-planar with the maroon circle. }
\label{fig:proof2}
\end{figure}

\begin{figure*}[t]
    \centering
    \makebox[\linewidth][c]{
        \setlength{\fboxsep}{0pt}
        \setlength{\fboxrule}{0.5pt}
        
        \newcommand{\mygraph}[2]{
            \begin{tikzpicture}[spy using outlines={rectangle,yellow,magnification=2,width=1.7cm,height=1.7cm}]
              \node {\includegraphics[height=0.25\linewidth,width=0.3\linewidth,trim=120mm 70mm 70mm 0mm, clip=true]{#1}};
              \spy on (0.45,-0.8) in node [left] at (2.6,0.8);
              \node[yshift=0.11\linewidth] {\color{white} #2};
            \end{tikzpicture}
        }

        \begin{tabular}{c|cc}
            tree reconstruction (blue) on  & 
            \multicolumn{2}{c}{tree reconstructions (red and green) on} \\ 
            {\bf undirected}\! Tubular\! Graph\! (standard) & 
            \multicolumn{2}{c}{{\bf directed} Confluent Tubular Graph (our)} \\ \hline
            \mygraph{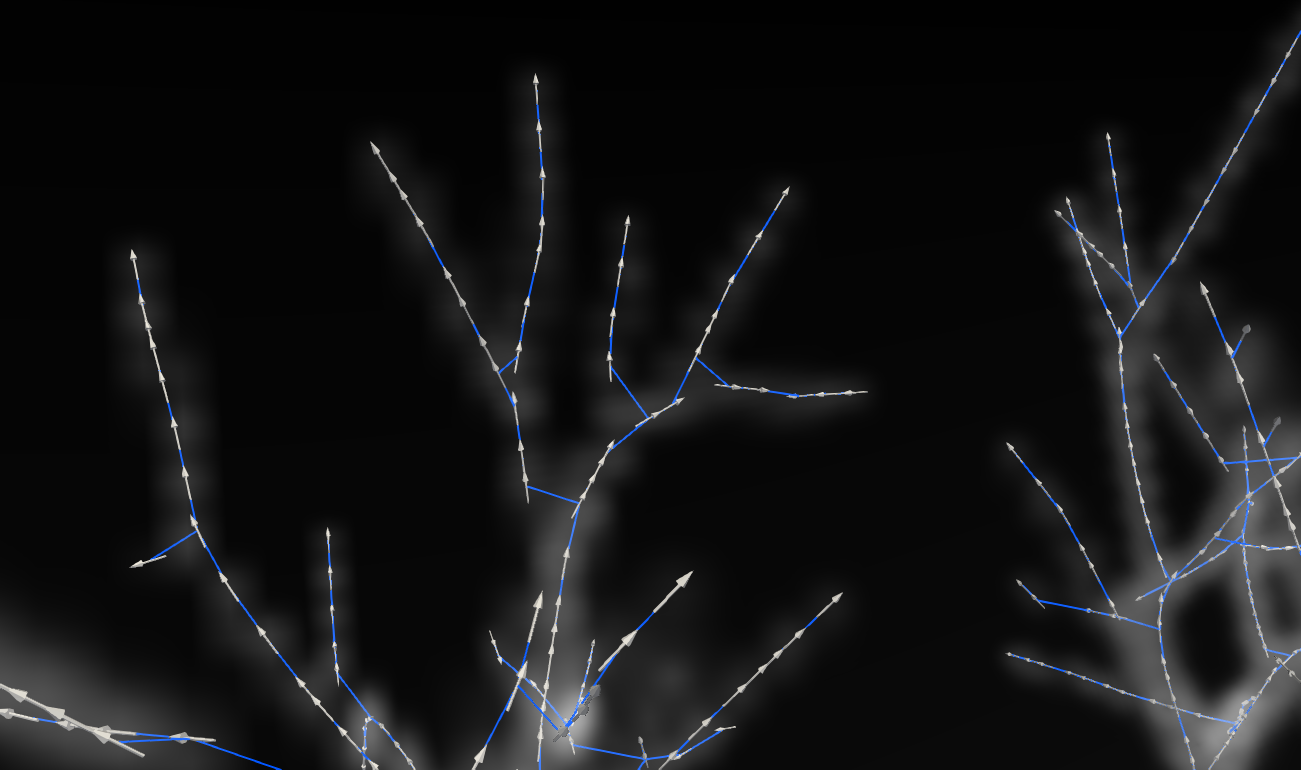}{GridMST} & \mygraph{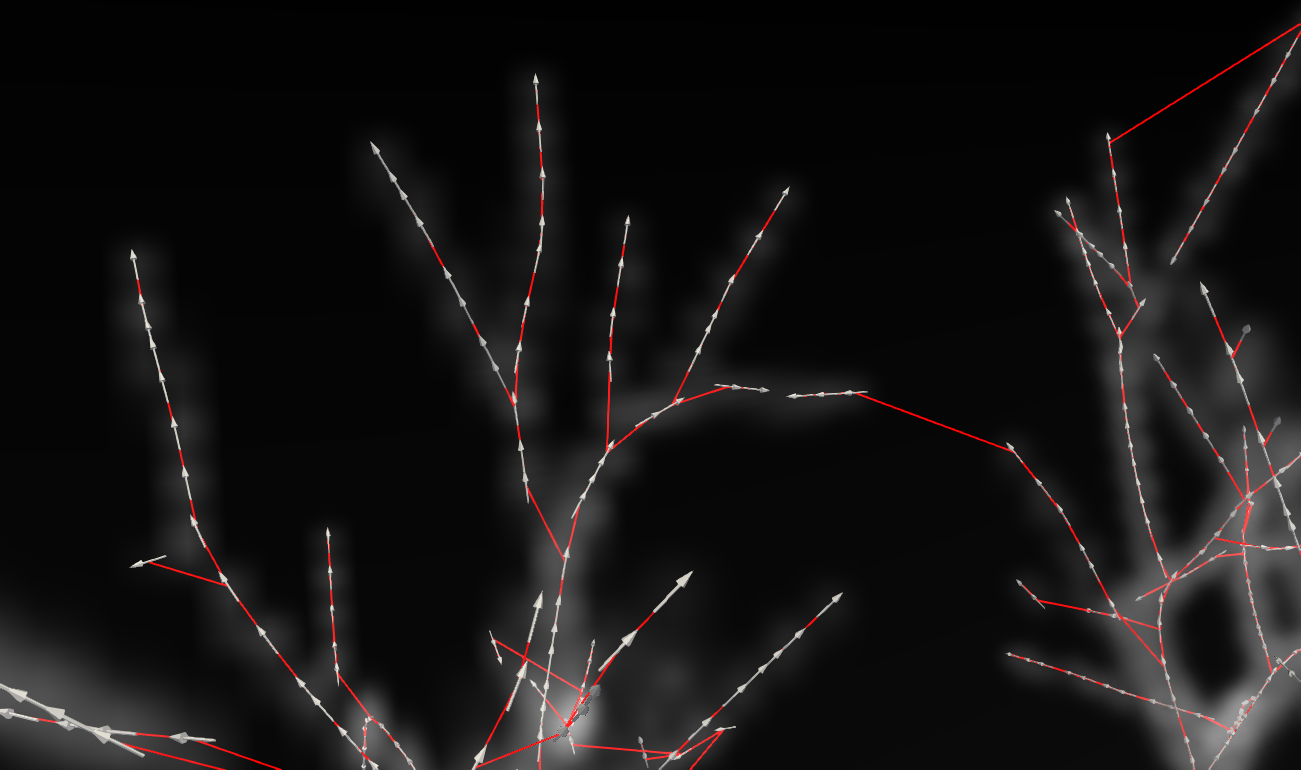}{GridArb} & \mygraph{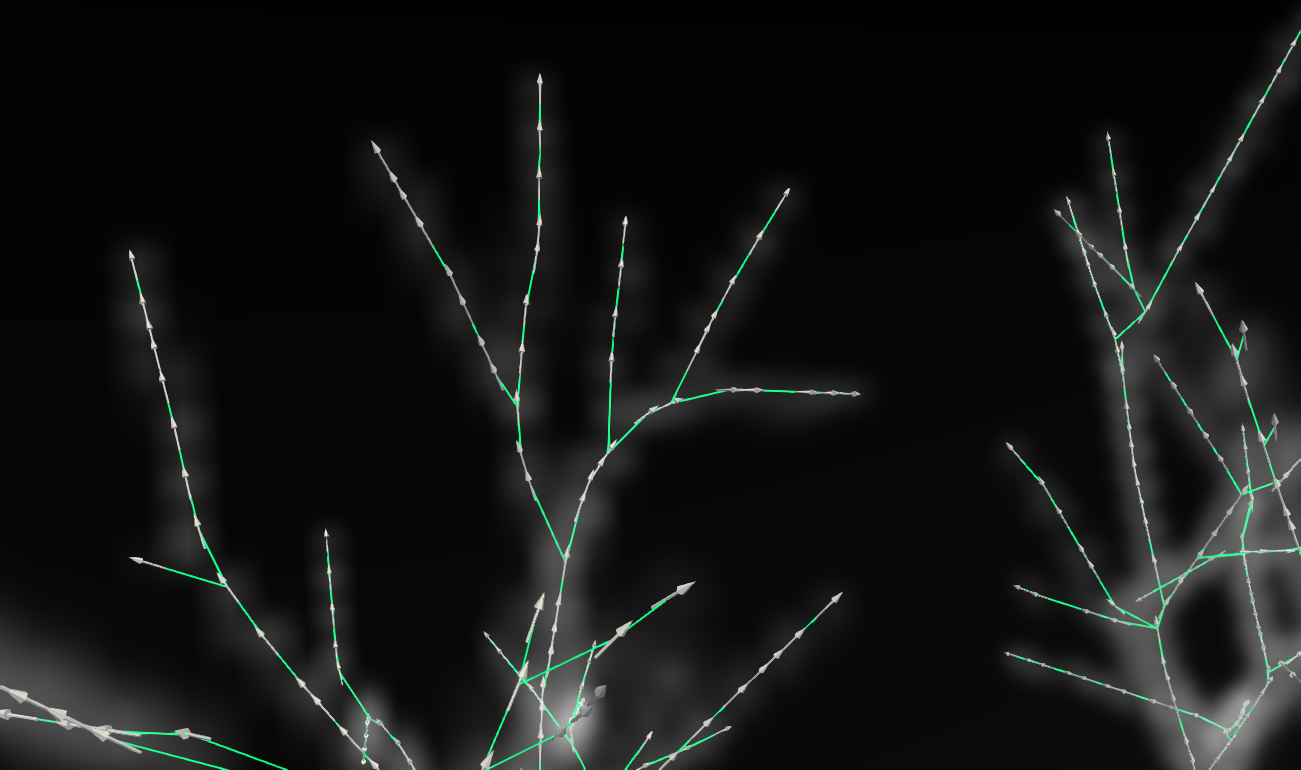}{MArb} \\
            \scriptsize (a) Flow pattern estimate (white) \cite{Divergence:cvpr19} & \scriptsize (b) Flow pattern estimate (white) \cite{Divergence:cvpr19} & \scriptsize (c) \emph{Improved} flow estimate (white)  \\
            \scriptsize + MST on geodesic tubular graph & \scriptsize + min. arb. on confluent tubular graph & \scriptsize + min. arb. on confluent tubular graph
        \end{tabular}
    }
    \caption{Typical tree reconstruction examples for standard (a) and our {\em confluent} (b,c) tubular graphs. (a) White vectors represent CRF-based flow pattern estimates \cite{Divergence:cvpr19} using 26-grid regularization 
    neighborhood $\cal N$. In case of thin sub-voxel vessels, such $\cal N$ has gaps creating inconsistent flow pattern for isolated small branches (yellow box) lacking bifurcations used by divergence prior to disambiguate orientations.
    (a) Undirected geodesic tubular graph with large KNN easily bridges such gaps ignoring (inconsistent) flow directions 
    and produces topologically valid vessel MST (blue), even though bifurcations are not accurate. (b) Directed confluent tubular graph is sensitive to flow pattern errors. Minimum arborescence on this graph produces accurate bifurcations, 
    but flow orientation errors (yellow box) lead to wrong topology. (c) CRF-based flow pattern estimator \cite{Divergence:cvpr19} 
    with modified anisotropic KNN system $\cal N$ addresses the gaps at thin vessels. 
    This improves flow orientations (white vectors in yellow box) and resolves confluent tubular graph artifacts producing trees with 
    accurate both topology and bifurcations.
    }
    \label{fig:gridarb}
\end{figure*}


\begin{proof}
First, we only consider the circle going through $p$, $q$ and tangential to some vector $\bar \tau_p$ as shown in Fig.~\ref{fig:proof1}. Moving the tangent vector $\bar \tau_p$ in its direction along the circle yields another tangent vector $\bar \tau_q$ at $q$. We also translate $\bar\tau_p$ to $q$. By construction, $\Updelta Opq$ is equilateral. Thus, $\angle Opq=\angle Oqp$. Since $\bar \tau_p$ and $\bar \tau_q$ are tangential to the circle, we have $\alpha+\angle Opq=\beta+\angle Oqp=\frac{\pi}{2}$, which obviously gives $\alpha=\beta$. WLOG, we assume vectors $\bar \tau_p$, $\bar \tau_q$ and $\bar e$ are all unit vectors. As $\alpha=\beta$, we have:
\begin{equation}
\label{eq: proof1}
\bar \tau_q=2[(\bar \tau_p\cdot\bar e)\bar e-\bar \tau_p] + \bar \tau_p
\end{equation}

Now, we consider the two circles going through both $p$ and $q$ while one is tangential to $\bar l_p$ and the other is tangential to the $\bar l_q$ as shown in Fig.~\ref{fig:proof2}. Note that these two circles are not necessarily co-planar. Using \eqref{eq: proof1}, we can obtain
\begin{equation}
\label{eq:sup cpq1}
\breve c_{pq}^1=2[(\breve c_{pq}^0\cdot \vec e_1)\vec e_1-\breve c_{pq}^0]+\breve c_{pq}^0
\end{equation}
\begin{equation}
\label{eq:sup cqp1}
\breve c_{qp}^1=2[(\breve c_{qp}^0\cdot \vec e_1)\vec e_1-\breve c_{qp}^0]+\breve c_{qp}^0
\end{equation}
To prove \eqref{eq:angle_identity}, it is sufficient to prove equality of the two dot products which can be simplified using \eqref{eq:sup cpq1} and \eqref{eq:sup cqp1}:
\begin{equation}
\label{eq: sup LHS}
\breve c_{pq}^1\cdot \breve c_{qp}^0=2(\breve c_{pq}^0\cdot \vec e_1)(\breve c_{qp}^0\cdot \vec e_1)-\breve c_{pq}^0\cdot \breve c_{qp}^0
\end{equation}

\begin{equation}
\label{eq: sup RHS}
\breve c_{qp}^1\cdot \breve c_{pq}^0=2(\breve c_{qp}^0\cdot \vec e_1)(\breve c_{pq}^0\cdot \vec e_1)-\breve c_{qp}^0\cdot \breve c_{pq}^0
\end{equation}
It is obvious that the RHS of \eqref{eq: sup LHS} and \eqref{eq: sup RHS} are equal. Therefore, these two angles are equal.
\end{proof}

\section{CRF Regularization Neighborhood} \label{app: ANN}

Our tree extraction method is based on a directed {\em confluent tubular graph} construction 
$G =\langle V,A \rangle$ presented in Sec.~3 of the paper.
We proposed an approach that builds confluent flow-extrapolating arcs $\curv c_{pq}$ for our graph 
from estimated {\em oriented} flow vectors $\{\bar l_p\,|\,p\in V \}$. Specific flow orientations can be computed 
from Frangi filter outputs using standard MRF/CRF regularization methods \cite{Divergence:cvpr19} 
enforcing divergence (or convergence) of the flow pattern. However, as mentioned in Sec.\ 3.3 and Sec.~4, we modified \cite{Divergence:cvpr19} by anisotropically enlarging the regularization neighborhood to improve the estimates of flow orientations, 
which are important for our directed arc construction. The 26-grid neighborhood regularization used 
in \cite{Divergence:cvpr19} generates too many CRF connectivity gaps near the vessel tree periphery where the signal gets weaker. Such gaps result in flow orientation errors, see white vectors in the zoom-ins in Fig.~\ref{fig:gridarb}(a,b). 
While tree reconstruction on standard \underline{undirected} geodesic tubular graphs, see Fig.~\ref{fig:gridarb}(a), 
are oblivious to such errors, our \underline{directed} {\em confluent tubular graph} construction is 
sensitive to wrong orientations, see Fig.~\ref{fig:gridarb}(b). 
To address CRF gaps in the flow orientation estimator \cite{Divergence:cvpr19}, 
we modified their 26-grid regularization neighborhood into \emph{anisotropic KNN} based on Frangi's vessel tangents
\cite{frangi1998multiscale}. This significantly reduces orientation errors in $\{\bar l_p\,|\,p\in V \}$
and resolves confluent tubular graph artifacts, 
see Fig.~\ref{fig:gridarb}(c). We detail anisotropic KNN below.


\paragraph{CRF connectivity quality:} Besides the size of the neighborhood $K$, anisotropic KNN system has another important hyper-parameter, aspect ratio $ar$. To select better parameters $K$ and $ar$, we can evaluate CRF connectivity system $\cal N$ using ROC curves for synthetic vasculature volumes with ground truth. We consider an edge in $\cal N$ as 
\emph{correct} iff the projections of its ends onto the ground truth tree have parent/descendant relationship. The \emph{recall} is the portion of the ground truth tree covered by the correct edges. The \emph{fall-out} is the ratio of incorrect edges to the total edge count.

\begin{figure}[t]
    \centering
    \begin{tikzpicture}[]{}
        \node[inner sep=0] at (0,0) {
            \includegraphics[width=0.9\linewidth]{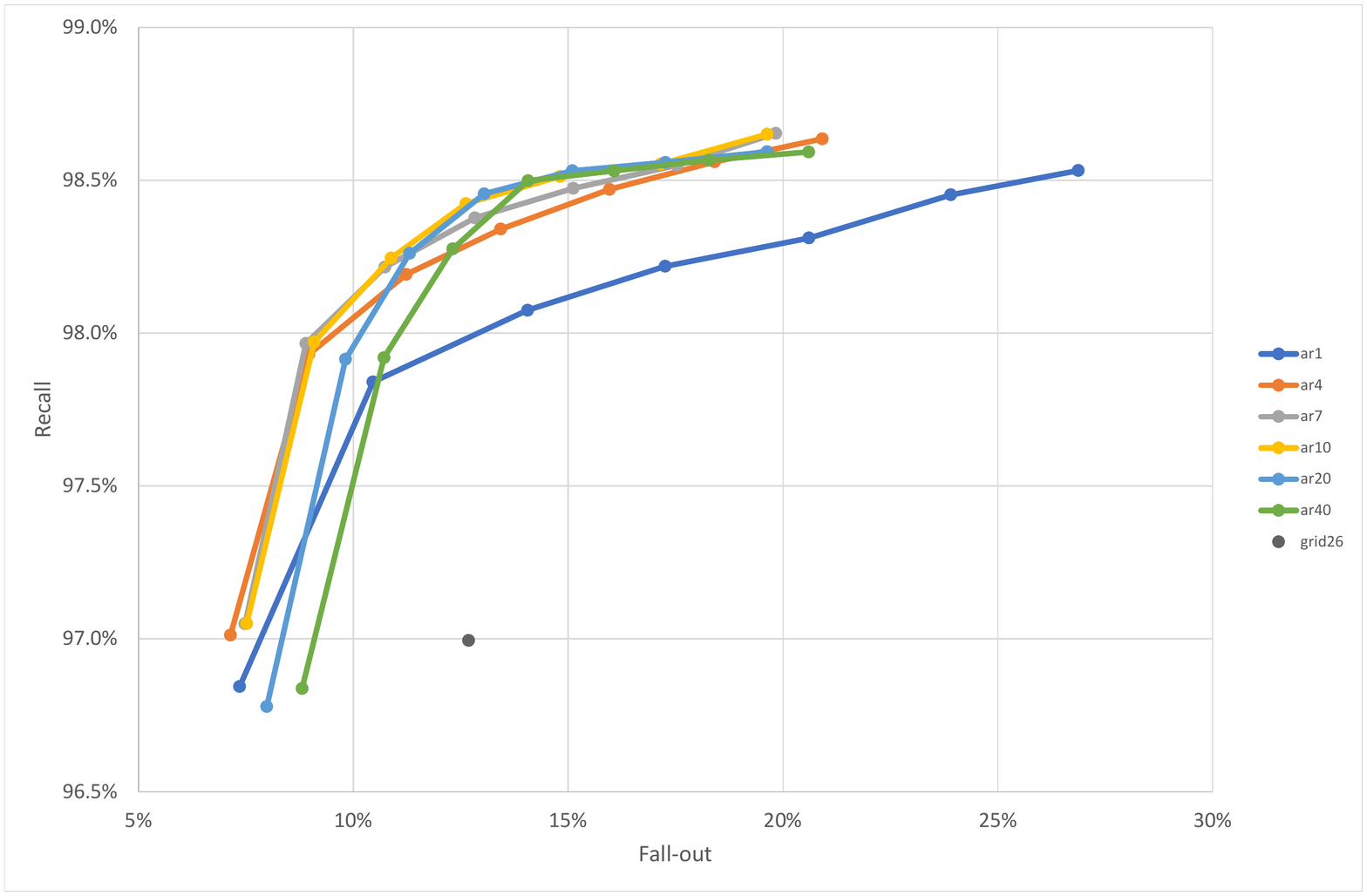}
        };
        
        \node (A) at (-0.7,-1.5) {\color{gray} $1$-ANN graph};
        \draw [gray] (A)+(-1.3,.5) circle (5mm);
        
        \node (A2) at (-0.4,0) {\color{gray} $2$-ANN graph};
        \draw [gray] (A2)+(-1.3,.5) circle (4mm);
        
        \node (A3) at (-2.2,1.7) {\color{gray} $4$-ANN graph};
        \draw [gray] (A3)+(1.3,-.5) circle (2.5mm);
        
        \node (A5) at (2.6,2.05) {\color{gray} $7$-ANN graph};
        \draw [gray] (A5)+(-1.3,-.5) circle (12mm and 2.3mm);
        
        \node (A4) at ($(A3)!0.5!(A5)$) {\color{gray} \dots};
    \end{tikzpicture}
    \caption{(Quasi) ROC curves evaluating accuracy of the neighborhoods $N$ used for flow pattern estimation, as in \cite{Divergence:cvpr19}. 
    We compare anisotropic KNNs and standard 26-grid connectivity (see gray dot). Evaluation is done based on synthetic data with ground truth where correct connectivity is available. ``ar" stands for the aspect ratio and the number denotes the square of the aspect ratio. ``grid26" represent the regular 26 neighbors on grid. We select ``ar10'' with 4 anisotropic nearest neighbourhood (ANN) connectivity system.}
    \label{fig:connectivity ROC}
\end{figure}

As shown in Fig.~\ref{fig:connectivity ROC}, simply increasing the size of neighborhood closes many gaps but, in the meantime, introduces a lot of spurious connections between different vessel branches. Thus, we propose to use anisotropic neighborhoods. Specifically, the regularization neighborhood is redefined as $k$ anisotropic nearest neighbors instead of regular grid connectivity. This similar to the KNN except Mahalanobis distance is used. This modification addresses the issue giving the state-of-the-art result, see Fig.\ref{fig:gridarb}(c). To implement such anisotropic neighborhood system, we first built an isotropic KNN with some large K, eg. K=500. Then, for each node and its neighbors, we transformed the Euclidean distance into Mahalanobis distance based on the tangent direction on the node. After this, we selected K (eg. K=4) nearest neighbors for each node again based on the Mahalnobis distance. Note that such anisotropic neighborhood is symmetric since we consider a pair of nodes as neighbor as long as one is connected to the other.
\section{Angular Error Measure} \label{app:anlge error}
The \emph{average} angular error introduced in \cite{Divergence:cvpr19} uses only correctly detected points to compute the bifurcation angular errors. Using such matching to compare different methods is unfair as for a particular detection threshold these methods correctly detect different sets of bifurcations. So, we match \emph{all} ground truth bifurcations to closet branching points on the detected tree regardless of their proximity. For certain thresholds, this causes many incorrectly matched bifurcation and large errors. Despite that such statistic is influenced significantly by random matches, it is meaningful for comparing different reconstruction methods.

\section{Synthetic Data with Ground Truth} \label{app:synthetic data}
 Zhang \etal \cite{Divergence:cvpr19} generated and published a dataset with ground truth using \cite{hamarneh2010vascusynth}. We found that the diversity of bifurcation angles is limited. The mean angle is $68^{\circ}$ and \textit{std} is $17^{\circ}$. To increase the angle variance, we introduce a simple modification of vessel tree generation. When a new bifurcation is created from a point and existing line segment, we move the bifurcation towards one of the segment's ends chosen at random decreasing the distance by half. The new mean is $68^{\circ}$ and \textit{std} is $29^{\circ}$. We generate 15 volumes $100\!\times\!100\!\times\!100$ with intensities between $0$ and $512$. The voxel size is $0.046$ mm. We add Gaussian noise with {\em std} 10 and 15. 

\end{document}